\documentclass[lettersize,journal]{IEEEtran}
\usepackage{amsmath,amsfonts}
\usepackage{algorithmic}
\usepackage{algorithm}
\usepackage{array}
\usepackage[caption=false,font=normalsize,labelfont=sf,textfont=sf]{subfig}
\usepackage{textcomp}
\usepackage{stfloats}
\usepackage{url}
\usepackage{xcolor}
\usepackage{verbatim}
\usepackage{graphicx}
\usepackage{cite}
\begin{document}

\title{SkeFi: Cross-Modal Knowledge Transfer for Wireless Skeleton-Based Action Recognition}

\author{Shunyu Huang, Yunjiao Zhou, Jianfei Yang,~\IEEEmembership{Senior Member,~IEEE}
\thanks{S. Huang and Y. Zhou are with the School of Electrical and Electronics Engineering, Nanyang Technological University, Singapore (e220035@e.ntu.edu.sg; yunjiao001@e.ntu.edu.sg).}

\thanks{J. Yang is with the School of Mechanical and Aerospace Engineering and jointly with the School of Electrical and Electronics Engineering at Nanyang Technological University, Singapore. J. Yang is the corresponding author (jianfei.yang@ntu.edu.sg).}}

\markboth{}%
{Shell \MakeLowercase{\textit{et al.}}: A Sample Article Using IEEEtran.cls for IEEE Journals}


\maketitle
\begin{abstract}
   Skeleton-based action recognition leverages human pose keypoints to categorize human actions, which shows superior
generalization and interoperability compared to regular end-to-end action recognition. Existing solutions use RGB cameras to annotate skeletal keypoints, but their performance declines in dark environments and raises privacy concerns, limiting their use in smart homes and hospitals. This paper explores non-invasive wireless sensors, i.e., LiDAR and mmWave, to mitigate these challenges as a feasible alternative. Two problems are addressed: (1) insufficient data on wireless sensor modality to train an accurate skeleton estimation model, and (2) skeletal keypoints derived from wireless sensors are noisier than RGB, causing great difficulties for subsequent action recognition models. Our work, SkeFi, overcomes these gaps through a novel cross-modal knowledge transfer method acquired from the data-rich RGB modality. We propose the enhanced Temporal Correlation Adaptive Graph Convolution (TC-AGC) with frame interactive enhancement to overcome the noise from missing or inconsecutive frames. Additionally, our research underscores the effectiveness of enhancing multiscale temporal modeling through dual temporal convolution. By integrating TC-AGC with temporal modeling for cross-modal transfer, our framework can extract accurate poses and actions from noisy wireless sensors. Experiments demonstrate that SkeFi realizes state-of-the-art performances on mmWave and LiDAR. The code is available at \url{https://github.com/Huang0035/Skefi}.
\end{abstract}


\section{Introduction}
\label{sec:intro}

\IEEEPARstart{A}{ction} recognition is crucial in mobile systems and ubiquitous computing, mainly utilizing video data (RGB image sequences) from conventional cameras to categorize human actions without specialized sensors, enhancing its versatility. Among action recognition studies, skeleton-based action recognition (SAR) has emerged as a robust solution~\cite{RNN, CNN, ST-GCN}, which uses deep learning to analyze skeletal joint positions and movements to discern and classify actions, boasting resilience to changes in lighting, background complexity, and visual heterogeneity. Consequently, action recognition has become increasingly significant. Currently, the vision-based solution is a predominant method in SAR based on deep learning methods, e.g., Openpose~\cite{openpose}, Nuitrack~\cite{Nuitrack}, and PoseNet~\cite{posenet}. These approaches have been widely used in many multimedia fields, including human-computer interaction~\cite{hr}, virtual reality~\cite{VR}, medical care~\cite{medical}, etc.

\begin{figure}[t]
\centering
\includegraphics[width=0.485\textwidth]{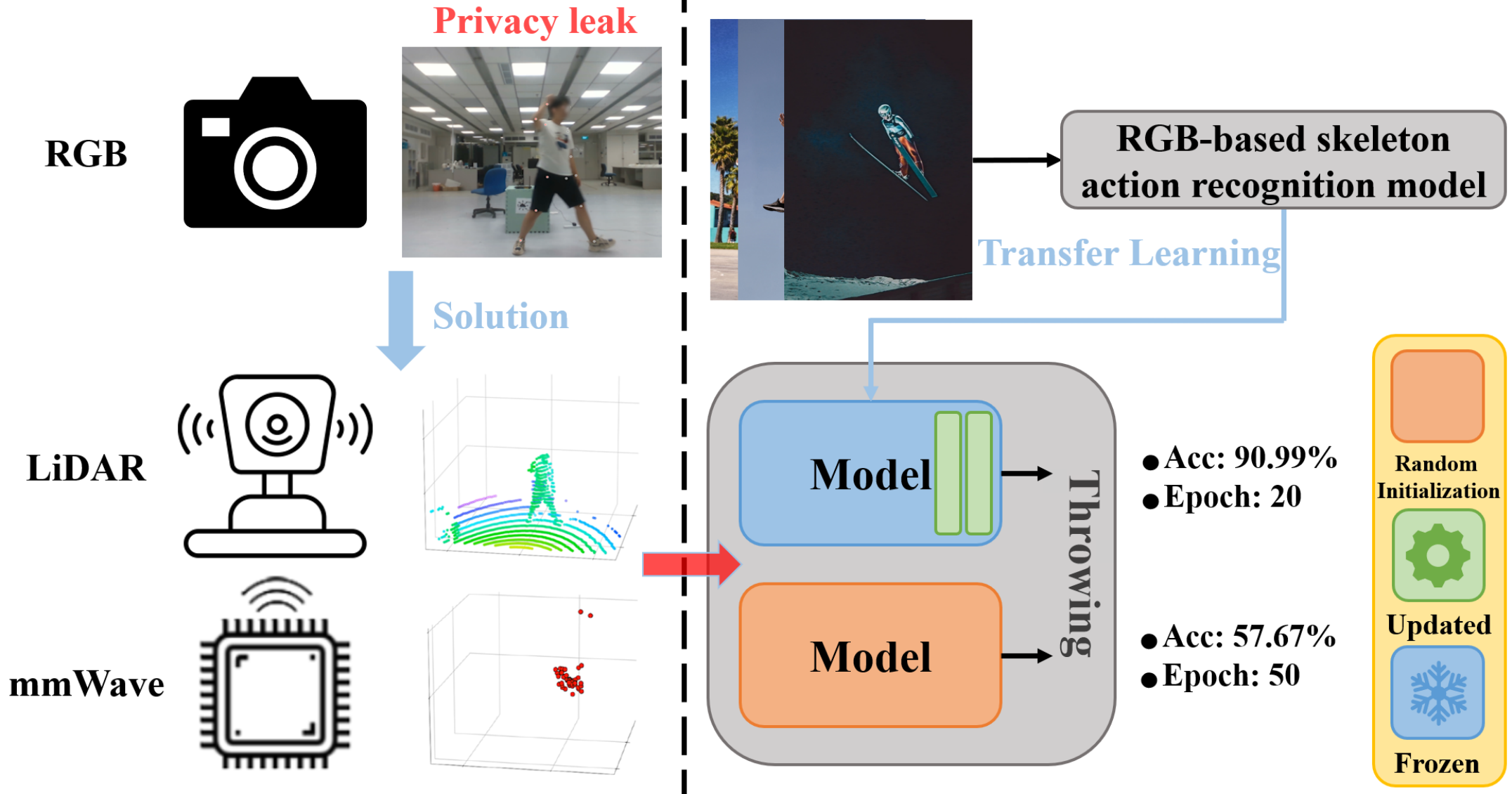}
\caption{SkeFi uses non-invasive wireless sensors to recognize human actions through skeletal movements. It deals with the lack of wireless sensor data by cross-modal transfer learning.}
\label{fig1}
\end{figure}

The deployment of action recognition systems using RGB cameras often raises privacy concerns due to the potential for unintentional collection of personal data without explicit consent in applications such as city surveillance, smart homes, and autonomous vehicles. For example, monitoring pedestrian flow inadvertently captures individual facial details, and in smart homes, the daily activities of residents could be recorded. Such data could be misused or leaked without strong privacy protections, endangering user privacy and safety. Additionally, factors like lighting, environmental backgrounds, and video resolution can affect the accuracy of keypoints annotation, impacting the practical use of SAR in real scenarios. In response to these challenges, recent advancements in wireless sensing technology~\cite{yang2023sensefi}, exemplified by innovations such as LiDAR~\cite{Lidar}, mmWave~\cite{mmWave}, and WiFi~\cite{wifi}, have gained prominence. These technologies utilize non-invasive sensors that capitalize on the spatial propagation properties of laser pulses or electromagnetic waves to detect and monitor human motion and location, offering viable alternatives to traditional vision-based solutions and addressing issues related to lighting conditions, privacy, and user convenience~\cite{nirmal2021deep}.

While non-invasive sensing technology holds promise, it introduces several inherent challenges to achieving robust SAR: (1) Non-intrusive measurement equipment is generally more costly than the widely available and resource-abundant RGB equipment. Moreover, the application scenarios for such technologies are not yet widespread, leading to a scarcity of data. (2) The limited data availability can make complex deep-learning models prone to overfitting. These models require a diverse dataset to train robustly. Still, with restricted data, they may overemphasize atypical samples that do not represent the general trend, thus diminishing the model's ability to generalize. (3) Skeletal keypoints derived from non-invasive sensors are noisier than those obtained from RGB modalities. In the RGB domain, specialized deep learning tools~\cite{densepose} for keypoint detection are tailored for precision. In contrast, non-invasive pose estimation methods often employ generic deep learning architectures like ResNet-48~\cite{Resnet}, which may produce noisier results in each frame. Additionally, these sensors are susceptible to frame loss due to inherent hardware limitations~\cite{mmWave}, compromising the accuracy and reliability of SAR tasks based on wireless sensing.

To overcome the aforementioned challenges, we propose cross-modal transfer learning to address the scarcity of data and enhance feature learning in the presence of missing frames. Transfer learning~\cite{transfer, transfer1} has proven an effective strategy for preventing overfitting in deep learning models trained on limited datasets. The field of skeletal motion recognition, traditionally reliant on the RGB modality, offers expansive datasets and extensive pre-existing knowledge. By leveraging this knowledge, we can enrich our model with latent patterns from the RGB domain, thus introducing vital prior knowledge into the tasks of non-invasive action recognition. Furthermore, to combat the issues of noise and frame loss inherent in skeletal keypoint estimation, we have adopted a strategy that involves integrating structural correlations from the spatial domain into the temporal domain. Introducing temporal correlations allows for capturing joint correlations across frames, effectively addressing the frame loss issue. Consequently, we have enhanced the model's capacity to process temporal information by incorporating a temporal correlation module, significantly boosting its feature learning capabilities.

This paper proposes SkeFi, a cross-modal knowledge transfer framework with a skeleton-based GCN backbone, designed to transfer knowledge from the RGB-based Kinetics-Skeleton dataset to the non-invasive dataset. SkeFi applies to all GCN models and leverages AGCN~\cite{AGCN} as its foundation, incorporating spatial and temporal modeling. To enhance the feature extraction capabilities of non-invasive data in cross-modal knowledge learning, we incorporate dual temporal convolution within the spatial modeling, enabling the nuanced capture of temporal information at individual keypoints. Furthermore, inspired by ESPNet~\cite{espnet}, we have enhanced multi-scale temporal modeling (MST) by employing additive operations, thus augmenting model efficacy without the necessity for additional computational overhead, a method we designate as ESP-MST. SkeFi demonstrates strong performance across novel non-invasive dataset MM-Fi's~\cite{MM-Fi} RGB, mmWave, and LiDAR modalities, achieving effective cross-modal knowledge learning. The overall architecture of SkeFi is illustrated in Figure~\ref{fig1}.

Our contributions are summarized as follows:
\begin{itemize}
    \item We propose SkeFi, a novel SAR solution using wireless sensors to address privacy violations and environmental limitations arising from RGB modality. As far as we know, it is the first work that leverages non-intrusive sensors and cross-modal learning for robust SAR.
    \item To deal with the data shortage in wireless sensors, we propose to transfer the insights gained from RGB-based skeleton action recognition to a non-intrusive mode to enhance the model's generalization ability. 
    \item We propose the TC-AGCN model to address data frame loss and noise in non-invasive data. Moreover, our refined ESP-MST not only bolsters the accuracy of the existing model but also augments the accuracy after transfer learning.
    \item Extensive experiments are conducted on a multi-modal SAR dataset, demonstrating that SkeFi achieves robust and good performances.
\end{itemize}

\section{Related work}
\label{sec:format}

\subsection{Skeleton Action Recognition}
Traditional methods for skeleton-based action recognition predominantly utilize hand-crafted features and joint relationships. Such approaches frequently encounter performance limitations, primarily because they fail to focus on the essential semantic linkages intrinsic to the human body. As deep learning has progressed, data-driven methodologies have garnered heightened interest, with RNNs and CNNs emerging as the principal models of choice. Nevertheless, neither RNN-based nor CNN-based approaches wholly capture the unique structure of skeleton data, given that the skeleton data diverges from a mere sequence of vectors or a 2D grid and is fundamentally embodied within a graph structure~\cite{RNN, RNN1, CNN}.

Many GCN (Graph Convolutional Networks) methods have recently been introduced to recognize human actions from skeleton data. GCNs are adept at capturing embedded features within irregularly structured data, presenting an advantage over RNN-based and CNN-based techniques. Specifically, GCNs eliminate the necessity for manual partitioning and traversal rules, leading to superior performance compared to earlier methods. The Spatio-Temporal Graph Convolutional Network (ST-GCN) has pioneered the domain of skeleton action recognition based on GCN~\cite{ST-GCN}. It directly takes the body skeleton sequence as input and extends the GCN into the spatiotemporal domain, thereby more effectively extracting the motion characteristics of actions.
Additionally, it introduces a novel weight allocation strategy, allowing for more differentiated learning of features across various nodes. Building upon the foundation of ST-GCN, the 2s-AGCN introduces graph adaptability with self-attention and a learnable graph residual mask~\cite{AGCN}. It employs a dual-stream framework to utilize joint and bone features, enhancing performance simultaneously. MS-G3D introduces multi-scale convolution and spatiotemporal graph convolution operators for enhanced feature extraction to address the biased weighting issue and adeptly capture intricate spatiotemporal relationships~\cite{MSG3D}. Moreover, MS-G3D refines the temporal module by employing dilated convolutions at varied expansion rates, broadening the receptive field. CTR-GCN introduces a channel topology refinement network that dynamically learns varying topologies and adeptly aggregates joint features across distinct channels~\cite{CTRGCN}, thereby enhancing the GCN's capability to encapsulate topological information. DG-STGCN introduces a novel framework for SAR~\cite{DGSTGCN}. Regarding spatial modeling, it leverages a learned affinity matrix to discern the dynamic graph structure, moving away from traditional fixed graph structures. It integrates a dynamic joint skeleton fusion module for temporal modeling to realize adaptive multi-point merging. While this approach has rendered promising outcomes, GCN-oriented techniques remain constrained, particularly with respect to robustness, compatibility, and scalability. Addressing these limitations, Duan et al. presented PoseC3D~\cite{PoseC3D}, a groundbreaking method for SAR. Distinctively, PoseC3D employs 3D heatmap stacking, eschewing graph sequences, as the foundational representation of the human skeleton. This pose estimation method is more resilient to noise and showcases enhanced generalization in cross-dataset scenarios. Consequently, PoseC3D has marked substantial performance augmentation across primary datasets~\cite{ki, NTU, finegym, ucf101}, both independently and when synergized with RGB modalities. However, most graph-based SAR methods are developed and evaluated on clean, continuous skeletons; when keypoints are noisy or frames are missing, their assumptions can break, motivating temporal-correlation modeling beyond per-frame graphs.

\subsection{Wireless Human Sensing}
With the advent of the Internet of Things (IoT), human sensing technologies have garnered significant interest, paving the way for innovative living scenarios. These encompass many applications, such as intelligent healthcare, security surveillance, and pervasive interactions. Scholars have delved into various sensing technologies to capture spatiotemporal and kinematic insights into human activities~\cite{WDHS1, WDHS2, WDHS3}. These sensing paradigms bifurcate into device-based and device-free categories. Device-centric methods employ wearable sensors~\cite{wearable1, wearable2}, ensuring versatile functionality across diverse settings. However, the necessity of donning these devices regularly can lead to potential inconvenience or occasional oversights.

In contrast, Wireless Device Human Sensing (WDHS) offers a non-intrusive alternative, leveraging tools like cameras~\cite{camera}, LiDAR, mmWave, and WiFi~\cite{wifi} to record human kinetics. Over the last decade, image-centric techniques have grown considerably, underpinning numerous commercial ventures. However, they grapple with inherent challenges, such as sensitivity to ambient illumination and potential privacy intrusions. Therefore, attention is turned to alternative sensors such as LiDAR~\cite{Lidar}, mmWave~\cite{mmWave, XRF55}, and WiFi~\cite{wifi2, wifi10}, which offer viable solutions~\cite{nirmal2021deep}. For instance, Li et al. exploited LiDAR's prowess in capturing precise depth data across diverse scenes~\cite{Lidar}, leading to the inception of the LiDARHuman26M dataset. This dataset heralded a novel research trajectory, emphasizing data-driven LiDAR-centric human motion capture over extended ranges. Concurrently, the evolution of millimeter-wave techniques has ushered in radio frequency (RF) imaging as a promising contender to offset wearable sensor constraints~\cite{mmWave2}. The mmWave radars output high-fidelity 3D point clouds, facilitating on-site processing via edge AI methodologies to delineate human motion~\cite{mmWave3, mmWave4, wang2023human}. Building on this, An et al.~\cite{mmWave} introduced a mmWave-based assistive rehabilitation system (MARS) tailored for smart healthcare, capable of 3D rendering of 19 distinct joints and skeletal structures. 

WiFi-based sensing, mainly through Channel State Information (CSI), has also garnered attention, primarily in gesture and action detection~\cite{zou2019wifi, gesture, wu2020fingerdraw}. Jiang et al. introduced WiPose~\cite{WiPose}, leveraging prior knowledge of the human skeleton in pose construction with RNN and smoothing the loss for fluid motion, but it is restricted to static positions. Ren et al. then developed Winect~\cite{Winect}, employing enhanced 2D AoA, signal separation, and keypoint modeling for dynamic activity tracking. The following year, Ren et al. proposed GoPose to boost 2D AoA spatial resolution by integrating the spatial diversity of the transmitter and Wi-Fi OFDM subcarriers' frequency diversity~\cite{GoPose}. 

Recently, Yang et al. unveiled MM-Fi, an innovative non-intrusive multi-modal 4D human pose dataset tailored for wireless human sensing~\cite{MM-Fi}. This dataset, which encapsulates the main wireless sensing mechanisms mentioned above, is the most holistic benchmark for wireless human posture estimation. It plays a pivotal role in spurring multi-modal human posture research and aids scholars in judiciously selecting sensor configurations based on specific requirements. Compared to RGB, wireless sensing pipelines typically provide noisier skeletons and limited training data, which motivates us to transfer priors from data-rich RGB and design a transfer learning architecture that is robust to noise and frame loss.

\subsection{Multimodal Transfer Learning And Knowledge Distillation}
{\color{black}Recent advances in cross-modal transfer for human activity recognition (HAR) have explored aligning heterogeneous sensing modalities through temporal synchronization and representation matching. Unsupervised Modality Adaptation frameworks such as C3T~\cite{c3t} align multi-sensor embeddings with contrastive objectives, while synchronized corpora like MM-Fi~\cite{MM-Fi} have enabled rigorous benchmarks for evaluating modality adaptation. In parallel, cross-modal knowledge distillation focuses on transferring supervision from strong modalities to weaker ones via shared embedding spaces and modality-specific distillation losses. RadarDistill~\cite{radardistill} distills spatial features from RGB videos into radar point clouds through multi-level alignment, and CRKD~\cite{crkd} introduces cross-resolution contrastive distillation between radar signals and visual embeddings. Compared with these approaches, SkeFi performs transfer at the skeleton representation level, avoiding the need to directly align high-dimensional radar maps or point clouds. Instead of employing explicit distillation losses or projection heads, SkeFi adopts a parameter-efficient fine-tuning strategy: early graph convolution blocks—encoding modality-agnostic joint structures—are frozen, while deeper temporal modules are adapted to address the noise, sparsity, and frame loss characteristic of LiDAR and mmWave skeleton sequences. This design enables lightweight and robust cross-modal transfer without introducing additional loss functions.

Building on these insights, Section III describes how SkeFi leverages an RGB-pretrained graph backbone together with adaptive temporal modules to achieve robust transfer across wireless skeleton modalities.}

\section{SkeFi framework}
\label{sec:pagestyle}

\subsection{Overview}
SkeFi is a transfer learning framework for SAR using GCNs. The framework incorporates spatial and temporal convolution modules to capture the structural information inherent in skeletal data. This section introduces relevant notation, defines standard graph convolutions, and then classifies and discusses current GCN algorithms by module. We finally elaborate on our improved model and detail the implementation of transfer learning within the SkeFi framework.

\subsection{Problem formulation}
The standard graph convolution employs the weight W for the feature transformation. It aggregates representations from the neighboring vertices of $v_{tj}$ via $a_{ij}$ to update the corresponding representation $f_{out}(v_{ti})$ of node $v_{ti}$, where $\left\{{v_{ti}}\mid t,i\in \mathbb{Z}, 1\le t\le T,1\le i\le N\right\}$ indicate $N$ body joints of the skeleton in a sequence of $T$ time steps. This process is formulated as:
\begin{equation}
    f_{out}(v_{ti})=\sum_{v_{tj}\in N(v_{ti})}^{} a_{ij}f_{in}(v_{ti})\mathrm{W}
\label{eq1}
\end{equation}
where $f_{in}(v_{ti})$ represents the input feature of node $v_{ti}$, and $a_{ij}$ is an element of the adjacency matrix $\mathrm{A}\in \mathbb{R}^{N\times N}$ capturing the graph's structure. $a_{ij}=1$ indicates a connection between nodes $v_{ti}$ and $v_{tj}$ at time step $t$, otherwise $a_{ij}=0$.

For static methods~\cite{ST-GCN}, the topology of the GCN remains unchanged during inference, with $a_{ij}$ typically defined manually. Conversely, dynamic methods allow the GCN's topology to be dynamically inferred during inference~\cite{AGCN, ASGCN}, with $a_{ij}$ generally determined by the model, contingent upon input samples.

\subsection{Spatial Modeling}
\noindent\textbf{ST-GCN~\cite{ST-GCN}.} In the spatial modeling module, the specific input feature map is a $C\times T\times N$ tensor, where $C$ represents the number of channels, $T$ represents the time length, and $N$ represents the number of vertices. To realize the ST-GCN, we transform Eq.~\ref{eq1} as follows:
\begin{equation}
    \mathrm{f}_{out}=\sum_{k}^{K_v}\mathrm{W}_k(\mathrm{f}_{in}\mathrm{A}_k)\odot \mathrm{M}_k
\label{eq2}
\end{equation}
Where $K_v$ denotes the kernel size of the spatial dimension, with the partition strategy designed by ST-GCN, $K_v$ is set to 3. $\mathrm{A}_k=\Lambda_{k}^{-\frac{1}{2}} \bar{\mathrm{A}}_k\Lambda_{k}^{-\frac{1}{2}}$, where $\mathrm{A}_k$ resembles an $N\times N$ adjacency matrix, wherein the element $\bar{\mathrm{A}}_{k}^{ij}$ denotes if the vertex $v_{j}$ belongs to the subset $N_i$. It extracts the connected vertexes in a particular subset from $\mathrm{f}_{in}$ for the corresponding weight vector. $\bar{\Lambda} _{k}^{ii}= {\textstyle \sum_{j}^{N}} \bar{\mathrm{A}}_{k}^{ij}+ \varepsilon$ is the normalized diagonal matrix with $\varepsilon = 0.001$ used to avoid empty rows. $\mathrm{M}_k \in \mathbb{R}^{N\times N}$ represents a learnable attention map that accentuates elements within each adjacency matrix. $\mathrm{W}_k \in \mathbb{R}^{C_{out}\times C_{in}\times 1\times 1}$ denotes weight matrix and $\odot$ denotes the dot product.

\noindent\textbf{2s-AGCN~\cite{AGCN}.} The GCN structure is hierarchical, encapsulating multi-layered semantic information across its layers. In contrast, ST-GCN employs a static approach, fixing the graph's topology across all layers, compromising its flexibility and capability to model the multi-layer semantic information inherent in each layer. To address this limitation, 2s-AGCN introduces an adaptive graph convolution layer. This approach optimizes the graph's topology and the network's other parameters in an end-to-end learning framework. To render the graph structure adaptive, 2s-AGCN modifies Eq.~\ref{eq2} as follows:
\begin{equation}
    \mathrm{f}_{out}=\sum_{k}^{K_v}\mathrm{W}_k\mathrm{f}_{in}(\mathrm{A}_k+\mathrm{B}_k+\mathrm{C}_k)
\label{eq3}
\end{equation}
where $\mathrm{A}_k$ is aligned with the original normalized $N\times N$ adjacency matrix $\mathrm{A}_k$ from Eq.~\ref{eq2}, signifying the physical structure of the human body. The second component, $\mathrm{B}_k$, is also an $N\times N$ adjacency matrix. Distinct from $\mathrm{A}_k$, the elements of $\mathrm{B}_k$ are parameterized and concurrently optimized with other network parameters during training. As $\mathrm{B}_k$ isn't constrained in its values, the resulting representation graph is learned solely from the training data. $\mathrm{C}_k$, represents a data-dependent graph that learns a distinct graph for every sample. To ascertain the existence and strength of a connection between two vertices, 2s-AGCN employs the normalized embedded Gaussian function to compute the similarity between them:
\begin{equation}
    f(v_i,v_j)=\frac{e^{\theta (v_i)^{T}\phi (v_j)}}{\sum_{j=1}^{N} e^{\theta (v_i)^{T}\phi (v_j)}} 
\label{eq4}
\end{equation}
where $N$ is the total number of the vertexes. In its specific implementation, 2s-AGCN employs the dot product to gauge the similarity between two vertices within the embedding space. Given an input feature map $\mathrm{f}_{in}$ of dimensions $C_{in} \times T\times N$, it undergoes embedding to dimensions $C_{e} \times T\times N$ using two $1 \times 1$ convolutional layers, designated as $\theta$ and $\phi$. The resultant embedded feature maps are reorganized and reshaped into $N\times C_{e}T$ and $C_{e}T\times N$ matrices. Their multiplication yields an $N\times N$ similarity matrix $\mathrm{C}_k$, where each element $C_{k}^{ij}$ reflects the similarity between vertex $v_i$ and vertex $v_j$. This matrix's values are then normalized to the range [0, 1].  $\mathrm{C}_k$ can be derived from Eq.~\ref{eq4} as follows:
\begin{equation}
    \mathrm{C}_k=softmax({\mathrm{f}_{in}}^T\mathrm{W}_{\theta k}^T\mathrm{W}_{\phi k}\mathrm{f}_{in})
\label{eq5}
\end{equation}
To maintain the original model's performance, 2s-AGCN doesn't substitute $\mathrm{A}_k$ with $\mathrm{B}_k$ or $\mathrm{C}_k$. Instead, it initializes both parameters and employs addition during training to bolster the model's adaptability.
\subsection{Spatial Modeling Improvements for Cross-Model Transfer}
\begin{figure*}[t]
\centering
\includegraphics[width=0.9\textwidth]{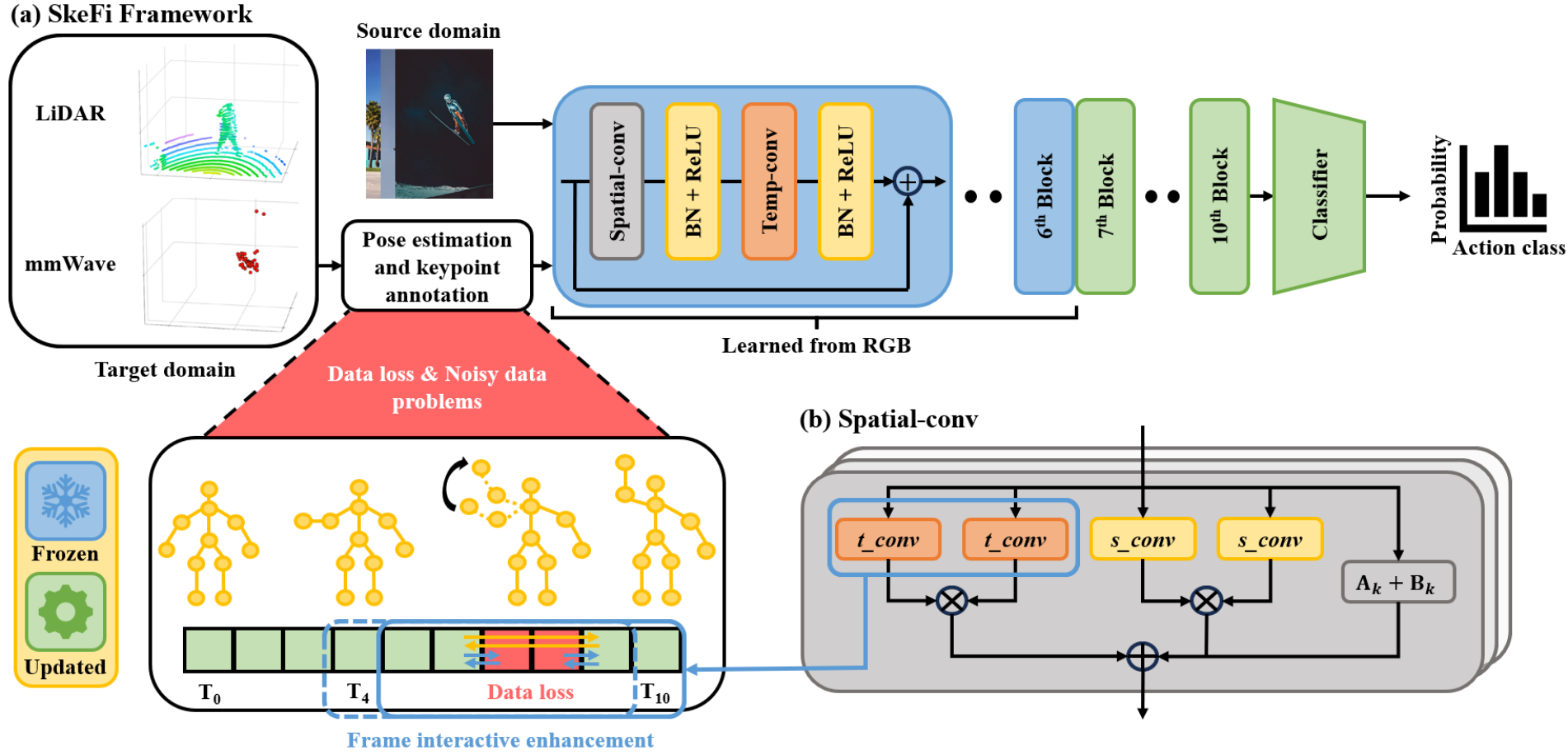}
\caption{(a). SkeFi, a transfer learning framework. It trains the model using RGB modality in the source domain and transfers prior knowledge to the target domain by freezing parameters. Green and blue denote trainable and frozen parameters, respectively. (b). TC-AGC. We propose a temporal correlation module, illustrated in the blue box, to boost connectivity in frame sequences by facilitating better interaction between missing and normal frames. $\bigoplus$ denotes the element-wise sum. $\bigotimes$ denotes the matrix multiplication.}
\label{fig2}
\end{figure*}

To mitigate the issues of noise and frame information loss after keypoints estimation from non-intrusive data, our approach incorporates the principle of structural correlation to improve the model. Structural correlations offer pivotal insights into the topology of graphs. However, in the cross-model knowledge transfer method proposed in this paper, existing models ignore the time correlation of source data, which may lead to the loss of potential patterns. Thus, We incorporate two temporal convolutions to enhance the interaction between missing and normal frames, improving the sequence's overall connectivity. The adaptive spatial modeling of temporal correlation (TC-AGC) is shown in Figure~\ref{fig2}(b). In the figure, frame loss at $T_7$ and $T_8$ in the target domain after skeleton estimation leads to incomplete hand-raising actions. Such loss in key action features could result in classification errors. Therefore, a time correlation module is introduced to maintain the action process's coherence and avoid misclassifying actions.

To incorporate temporal information into the adaptive spatial graphs, we construct $\mathrm{D}_k$ using the Gaussian function in Eq.~\ref{eq4}. $\mathrm{D}_k$ is analogous to $\mathrm{C}_k$ (Eq.~\ref{eq5}) in form - both use a bilinear similarity followed by a row-wise softmax - but differ in the instantiation of $\theta$ and $\phi$: $\mathrm{D}_k$ replaces $1 \times 1$ with $9 \times 1$ temporal convolutions with same padding along the time axis, which preserves the sequence length $\mathrm{T}$ and injects a local temporal inductive bias before affinity estimation. Consequently, the output feature in Eq. 3 becomes:

\begin{equation}
\mathrm{f}_{out}=\sum_{k}^{K_v}\mathrm{W}_k\mathrm{f}_{in}(\mathrm{A}_k+\mathrm{B}_k+\mathrm{C}_k+\mathrm{D}_k)
\label{eq6}
\end{equation}
\subsection{Temporal Modeling}
It is essential to disseminate information concerning the nodes' features across other skeletons in the sequence to capture the temporal dynamics within a skeleton sequence and consider the motions particular in action. Existing research can be categorized into single-scale temporal modeling and multi-scale temporal modeling.

\noindent\textbf{Single-Scale Temporal Modeling (SST). } In the case of single-scale temporal modeling, given that each vertex has a fixed number of neighbours (2, corresponding to joints in two consecutive frames), graph convolution can be employed akin to the conventional convolution operation. Precisely, a $K_t \times 1$ convolution is executed on the feature map derived from spatial modeling, where $K_t$ denotes the kernel size in the temporal dimension, typically set to 9.

\noindent\textbf{Multi-Scale Temporal Modeling (MST). } Multi-scale temporal modeling employs multiple parallel convolutional blocks to characterize the temporal relationships among joints in the input skeleton. This structure encompasses four branches: each integrates a $1 \times 1$ convolution to diminish the channel dimension. The initial two branches house two temporal convolutions with varying dilations, retaining a fixed kernel size of $3 \times 1$. The third branch incorporates a MaxPool operation, while the fourth branch utilizes a $1 \times 1$ convolution. The results from all four branches are concatenated to produce the output. This approach leverages varied dilation rates, eschewing the need for larger kernels to achieve an expansive receptive field. However, introducing dilated convolution can induce grid effects, compromising the continuity of the skeleton information. Drawing inspiration from the ESP module~\cite{espnet}, we integrate the temporal convolution outcomes from the first two branches, which utilize varying dilation rates before the ultimate connection. The improved method does not augment the complexity of the Multi-Scale Temporal Modeling and adeptly mitigates the grid effect. We abbreviate the improved Multi-Scale Temporal Modeling as ESP-MST.

\subsection{Cross-model Knowledge Transfer framework: SkeFi}
In our methodology, each joint's neighborhood spans the entire human skeleton graph, following prior work demonstrating the effectiveness of full-body receptive fields for action modeling~\cite{AGCN}. The complete network consists of ten graph convolution blocks, followed by global average pooling and a softmax classifier for final action prediction.

For cross-modal transfer, we adopt Kinetics-Skeleton~\cite{ki} as the source domain and MM-Fi~\cite{MM-Fi}—including its RGB, LiDAR, and mmWave modalities—as the target domain. This choice is motivated by three shared properties: (1) both datasets use an identical set of skeleton keypoints, eliminating the need for feature remapping; (2) both exhibit frame loss and sensor noise, allowing the robust spatiotemporal patterns learned from RGB skeletons to benefit wireless modalities; and (3) Kinetics-Skeleton provides large-scale supervision across 400 categories, enabling the model to learn modality-agnostic motion priors that are transferable to non-intrusive sensing tasks.

{\color{black}A core component of SkeFi is a block-wise layer-freezing strategy designed to preserve transferable spatial representations while enabling adaptation to target-domain noise characteristics. After initializing the network with weights pretrained on Kinetics-Skeleton, the first six graph convolution blocks—responsible for modality-agnostic structural encoding—are frozen. Only the deeper temporal blocks and the classifier are fine-tuned on MM-Fi. This selective adaptation prevents overfitting to the sparse and noisy LiDAR/mmWave skeletons while maintaining parameter efficiency.}

We adopt a two-stage training scheme with cross-entropy loss~\cite{ST-GCN, AGCN}. During the second stage, the fully connected classification head is replaced to match the number of target-domain categories. The complete transfer workflow is summarized in Algorithm~\ref{pseudo1}, and the adaptation procedure is illustrated in Fig.~\ref{fig2}(a).

\begin{algorithm}[ht]
\caption{{Cross-Modal Knowledge Transfer}}\label{pseudo1}
\begin{algorithmic}
\STATE
\STATE \textsc{TRAIN}(Kinetics-Skeleton features $f_{kin}$)
\STATE \hspace{0.5cm} Initialize spatial partition weights $\{\mathrm{W}_k\}$ and backbone (TC-AGC $\rightarrow$ Temporal);
\STATE \hspace{0.5cm} Set classifier to $N_s$ classes; mark all parameters trainable;
\STATE \hspace{0.5cm} \textbf{for} $epoch$ \textbf{do}
\STATE \hspace{1.0cm} $p_{kin} \leftarrow \textsc{Forward}(f_{kin})$;
\STATE \hspace{1.0cm} Compute cross-entropy loss $\mathcal{L}_{ce}$;
\STATE \hspace{1.0cm} Update all GCN/backbone and classifier parameters;
\STATE \hspace{0.5cm} \textbf{end for}
\STATE \hspace{0.5cm} Save pretrained weights $\theta^\star$;

\STATE
\STATE \textsc{SkeFi Framework}$(f_{mm}, \{\mathrm{W}_k\})$
\STATE \hspace{0.5cm} Load $\theta \leftarrow \theta^\star$; replace classifier with $N_t$ classes;
\STATE \hspace{0.5cm} Freeze shallow layers (early blocks); initialize/unfreeze deep layers and classifier;
\STATE \hspace{0.5cm} \textbf{for} $epoch$ \textbf{do}
\STATE \hspace{1.0cm} $x \leftarrow \textsc{AlignFrames}(f_{mm})$ \;\;(e.g., pad/sample to a fixed $T$);
\STATE \hspace{1.0cm} $p_{mm} \leftarrow \textsc{Forward}(x)$;
\STATE \hspace{1.0cm} Compute cross-entropy loss $\mathcal{L}_{ce}$;
\STATE \hspace{1.0cm} Update only unfrozen GCN/backbone layers and classifier;
\STATE \hspace{0.5cm} \textbf{end for}
\STATE \hspace{0.5cm} Return best checkpoint on validation;

\STATE
\STATE \textbf{Function} \textsc{Forward}$(x)$:
\STATE \hspace{0.5cm} \textbf{for each block} \textbf{do} $x \leftarrow \textsc{TC-AGC}(x)$; $x \leftarrow \textsc{Temporal}(x)$ \textbf{end for};
\STATE \hspace{0.5cm} \textbf{return} $\textsc{Softmax}(\textsc{FC}(\textsc{GAP}(x)))$;
\end{algorithmic}
\end{algorithm}

\section{Experiment}
\label{sec:typestyle}

We initially benchmarked the MM-Fi dataset using the standard training models. Next, we benchmark performance using the Kinetics-Skeleton dataset, followed by a transfer learning experiment transitioning from the Kinetics-Skeleton to the MM-Fi dataset with the SkeFi framework. We delve deeply into the variables influencing transfer learning results.

\subsection{Datasets}
\noindent\textbf{MM-Fi~\cite{MM-Fi}:} MM-Fi is a novel non-invasive multimodal 4D human pose dataset for wireless human sensing. Incorporating 1,080 sequential sequences, MM-Fi aggregates over 320k synchronized frames spanning five sensing modalities: RGB images, depth images, LiDAR point clouds, mmWave radar point clouds, and WiFi CSI data. This confluence of features distinguishes MM-Fi as the most exhaustive benchmark for wireless human body pose estimation currently accessible. The dataset is meticulously annotated, encompassing 2D/3D human pose landmarks, action classifications, 3D human spatial coordinates, and computed 3D dense poses. MM-Fi is the pioneering dataset offering five distinct non-invasive 4D human pose estimation (HPE) modalities.

\noindent\textbf{Kinetics-Skeleton~\cite{ki}:} Kinetics is a comprehensive human action dataset encompassing 300,000 video clips across 400 categories. These clips, sourced from YouTube, exhibit extensive diversity. Notably, the dataset offers only raw video clips and excludes skeletal data. The study in~\cite{ST-GCN} employed the OpenPose toolbox~\cite{openpose}, which is publicly accessible, to approximate the position of 18 joints for each segment in every frame. Within this dataset, each skeleton comprises 18 body joints. Every body joint is delineated by a 3D vector $(x,y,c)$, where $x$ and $y$ represent the 2D joint coordinates and $c$ stands for the confidence score. This configuration aligns with the NTU RGB+D dataset~\cite{NTU}, given that its input data manifests as a tensor of dimensions $3\times300\times18$. The dataset is partitioned into a training subset of 240,000 clips and a validation subset of 20,000 clips.

\subsection{Experiment Settings}
Experiments in this paper are executed using the PyTorch deep learning framework~\cite{pytorch}. For the Kinetics-Skeleton dataset, we adopt the input tensor size from~\cite{ST-GCN}, encompassing 150 frames with two objects each, and mirror the data augmentation method described in~\cite{ST-GCN}. Training employs stochastic gradient descent (SGD) with a Nesterov momentum of 0.9. We select cross-entropy loss for recognition, set a weight decay of 0.0001 for both search and training, use a batch size of 128, and adjust the learning rate of 0.1 by reducing it tenfold at the $45_{th}$ and $55_{th}$ epochs. The training process concludes at the $65_{th}$ epoch.

For the MM-Fi dataset's 2D and 3D human pose annotations, we initially employ the large-scale deep learning model HRNet-w48 to derive 2D keypoints~\cite{SunXLW19,HrNet}. The 17 keypoints generated adhere to the coco protocol~\cite{coco}. Subsequently, using the approach described in~\cite{MM-Fi}, we triangulate the two views of these 2D keypoints to procure the 3D keypoints ultimately. 

\begin{figure}[h]
\centering
\includegraphics[width=0.45\textwidth]{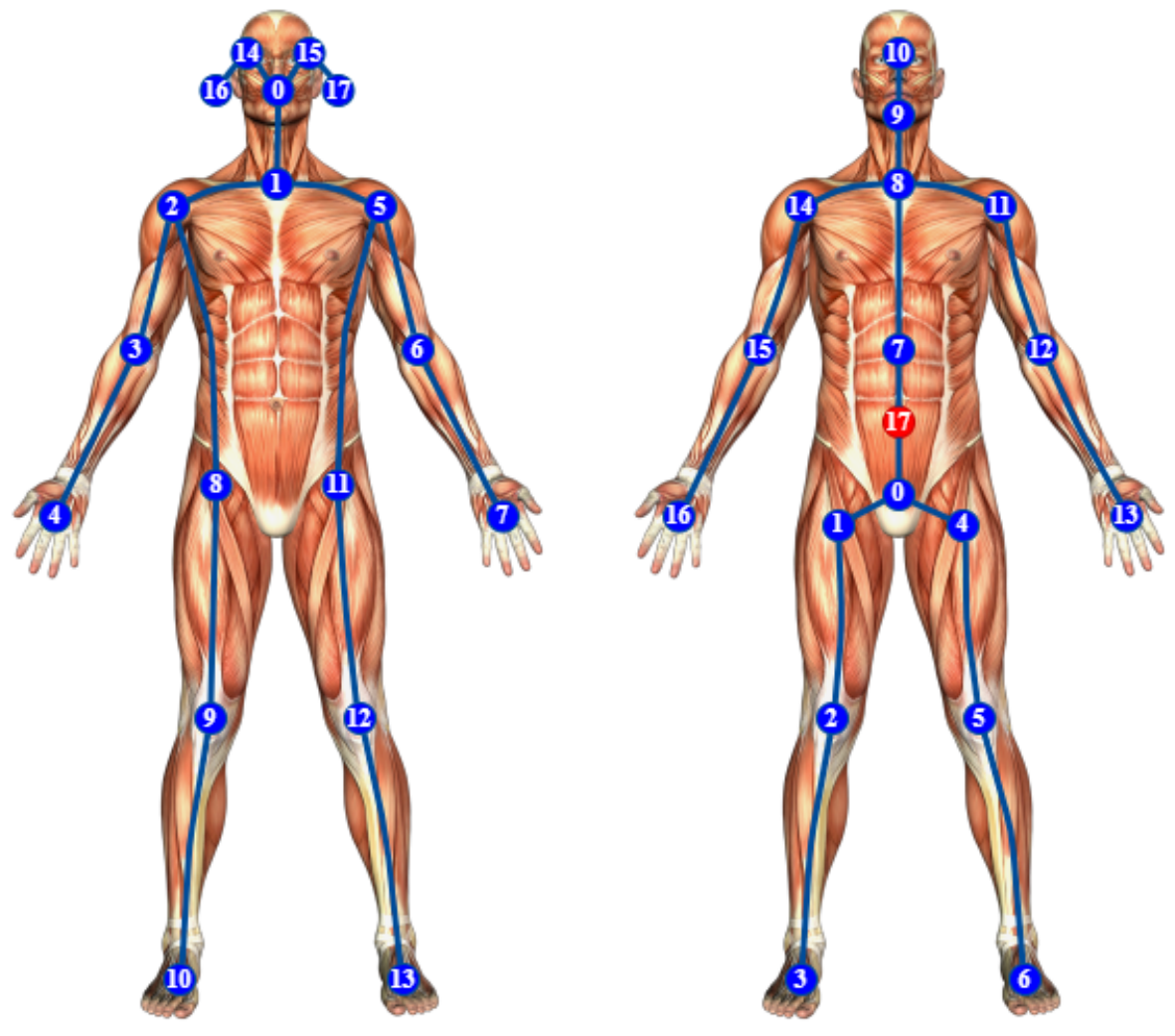}
\caption{The left sketch shows the joint label of the Kinetics-Skeleton dataset, and the right sketch shows the joint label extracted from the MM-Fi dataset.}
\label{fig3}
\end{figure}

\begin{figure*}[t]
\centering
\includegraphics[width=1.0\textwidth]{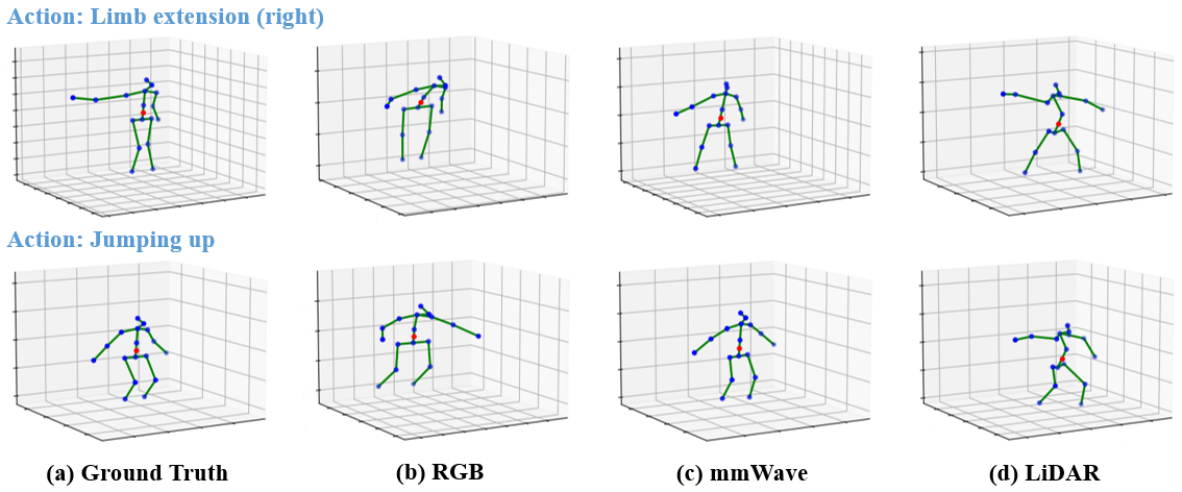}
\caption{Visualization of human pose estimation using three modalities. A red dot is keypoint we added.}
\label{fig4}
\end{figure*}

From the discerned 3D keypoints, we collected 270 samples, encompassing 27 actions by 
10 subjects. Each action tensor is sized at $3\times297\times17$. To align with the Kinetics-Skeleton input format, we undertook two steps: zero-padding the tensor's last three frames and introducing a $17_{th}$ point, determined as the midpoint between the $0_{th}$ and $7_{th}$ points (i.e., $\mathcal{V}_{17}=\left \{ \frac{{v_{0,t}}+{v_{7,t}}}{2}, 1\le t\le 300 \right \} $). This refines each action's tensor size to $3\times300\times18$. Unlike datasets such as NTU-RGBD, our 270 samples exhibit a distinct characteristic: a subject repeats the same action multiple times rather than executing varied actions. We segmented these repetitions and mirrored them to increase the number of samples. This yielded 5,400 samples, each with a tensor dimension of $3\times30\times18$. The joint positions of the Kinetics-Skeleton and our estimated 3D keypoints, along with their natural connections, are shown in Figure~\ref{fig3}. The red node indicates the added node. Figure~\ref{fig4} shows the visualization of the three modalities after data processing.

We first performed standard model training with seven subjects as the training set, while the other three subjects are designated as the validation set. A singular subject serves as the training set to validate the SkeFi framework, while the other nine are designated as the validation set. For standard training experiments on MM-Fi data, we use a batch size of 16. The learning rate is set as 0.05 and is divided by 10 at the $10_{th}$ and $20_{th}$ epoch. The training process is ended at $30_{th}$ epoch. For the transfer learning experiment, the learning rate is divided by ten at the $5_{th}$, $10_{th}$ and $15_{th}$ epoch. The training process is ended at $20_{th}$ epoch. Furthermore, we adopt the data augmentation approach from~\cite{ST-GCN} to enhance data diversity. Specifically, we zero-pad the input skeleton sequence to 40 frames and subtly alter joint coordinates using random rotations and translations.

\subsection{Ablation Study}
{\color{black}To evaluate the effect of different architectural components within SkeFi, we conduct ablation studies on the MM-Fi dataset across its three modalities: RGB, LiDAR, and mmWave. We focus our comparisons on GCN-based models because graph convolutional networks currently represent the most established and widely adopted backbone family for skeleton-based action recognition, offering explicit modeling of joint connectivity that aligns with the graph-based design of SkeFi. As a starting point, we include ST-GCN~\cite{ST-GCN} as a representative baseline, and further examine AGCN~\cite{AGCN} and the temporal-correlation variant TC-AGCN to assess the influence of stronger graph-based backbones. To analyze the contribution of temporal modeling, we additionally integrate the MST and ESP-MST modules into these architectures. This setup provides a consistent foundation for isolating the impact of each design choice, and the comparative results are summarized in Table~\ref{tb1}.}

\begin{table}[h]
    \caption{Classification accuracy comparison of state-of-the-art methods on three modalities of MM-Fi. Using a 7:3 ratio for the training and testing sets. The modules in parentheses represent the temporal modeling initially used by this model.}
    \label{tb1}
    \centering
    \begin{tabular}{lccc}
    \hline
    Methods                              & RGB (\%)                              & mmWave (\%)                           & LiDAR (\%)                                         \\ \hline
    STGCN (SST)~\cite{ST-GCN}                 & 85.97                                 & 63.70                                 & 45.37                                         \\
    AGCN (SST)~\cite{AGCN}                    & 92.72                                 & 72.62                                 & 71.27                                         \\
    AGC + MST~\cite{MSG3D}                    & 93.64                                 & 77.11                                 & 78.25  
                                  \\
    AGC + ESP-MST                             & 94.80                                 & \textbf{78.37}                        & 78.66  
                                  \\
    \hline
    TC-AGCN (SST)                             & 97.10                                 & 75.23                                 & 73.91  
                                  \\
    TC-AGC + MST                              & 98.45                                 & 78.01                                 & 78.49           
                                  \\
    \textbf{TC-AGC + ESP-MST}                    & \textbf{98.52} 
    & 72.98                                 & \textbf{79.21}  \\ \hline
\end{tabular}
\end{table}

Overall, while all models demonstrate strong performance on the RGB modality, they struggle with the non-intrusive modalities of mmWave and LiDAR, primarily due to noise and frame loss during keypoints estimation. The TC-AGC + ESP-MST method enhances accuracy, achieving 98.52\% in the RGB modality and 79.21\% in LiDAR. Although the TC-AGCN with ESP-MST performs slightly lower than AGCN in the mmWave modality, our innovative TC-AGCN approach surpasses ST-GCN and standard AGCN. This demonstrates that introducing temporal correlation modules can effectively improve performance on data with noise and frame loss. The introduction of MST can effectively improve model performance. Without changing the number of parameters, ESP-MST can further enhance performance, indicating that both MST and ESP-MST effectively increase the receptive field of the model, contributing to improved model performance.

\subsection{Transfer Learning to MM-Fi Dataset}
\noindent\textbf{Source/target choice and two-stage recipe.} We adopt Kinetics-Skeleton as the source and MM-Fi as the target because they share the same number of joints and both contain frame loss/noise, allowing source priors to benefit target learning while avoiding manual feature remapping. We first pretrain on Kinetics-Skeleton and then transfer to MM-Fi with layer freezing and a re-initialized classifier. The evaluation results of Kinetics-Skeleton are shown in Table~\ref{tb2}.
\begin{table}[h]
    \caption{Classification accuracy comparison with state-of-the-art
    methods on the Kinetics-Skeleton dataset.}
    \label{tb2}
    \centering
    \begin{tabular}{lcc}
    \hline
    Methods           & Top-1 (\%)          & Top-5 (\%)            \\ \hline
    STGCN~\cite{ST-GCN}       & 32.88               & 56.00                 \\ \hline
    AGCN~\cite{AGCN}     & 35.10               & 57.10                 \\ 
    AGC + MST~\cite{MSG3D}        & 35.73               & 58.31                 \\ 
    AGC + ESP-MST    & 35.80               & 58.41                 \\ 
    \hline
    TC-AGCN     & 35.59               & 58.19                 \\ 
    TC-AGC + MST     & 35.58               & 58.35                 \\ 
    \textbf{TC-AGC + ESP-MST} & 35.45               & \textbf{58.56}        \\ \hline
\end{tabular}
\end{table}

We observe that the incorporation of MST improves its performance by 0.63\% compared to the foundational AGCN. Without changing the number of parameters, ESP-MST integration elevates its performance by 0.7\%. While the introduction of MST and ESP-MST witnessed a marginal dip in the Top-1 accuracy of TC-AGCN, the Top-5 accuracy has further improved.

\noindent\textbf{Freezing policy.} 
The depth of fine-tuning is pivotal for transfer performance. Following the common view that early layers encode modality-agnostic skeleton geometry while later layers adapt to target specifics, we freeze \textbf{layers 1–6} and fine-tune layers 7–10 plus the classifier~\cite{freeze,freeze1}.
For clarity, we denote “L$k$–10” as the setting where only layers $k$–10 (and the classifier) are trainable; “All” indicates no freezing.
We evaluate this spectrum on AGCN+ESP-MST and TC-AGCN+ESP-MST, using the RGB modality as a representative example.
Table~\ref{tab:fig7_finetune_depth} shows a clear peak at \textbf{L7–10}, supporting our choice to freeze the first six layers during transfer while fine-tuning the higher blocks and classifier.

\begin{table}[h]
    \caption{{Fine-tuning depth vs.\ accuracy on the RGB modality.}}
    \label{tab:fig7_finetune_depth}
    \centering
    \setlength{\tabcolsep}{6pt}      
    \renewcommand{\arraystretch}{0.95} 
    \footnotesize
    \begin{tabular}{@{}lcc@{}}  
        \hline
        Fine-tuned layers & AGCN+ESP-MST (\%) & TC-AGCN+ESP-MST (\%) \\ \hline
        L10      & 80.78 & 81.52 \\
        L9--10   & 89.79 & 89.36 \\
        L8--10   & 90.76 & 91.85 \\
        \textbf{L7--10}   & \textbf{93.97} & \textbf{94.03} \\
        L6--10   & 92.16 & 92.45 \\
        L5--10   & 92.00 & 91.40 \\
        All      & 92.08 & 92.12 \\ \hline
    \end{tabular}
\end{table}

\noindent\textbf{Validation of SkeFi framework.} To assess the effectiveness of the transfer learning implemented in our study, we set the training-to-validation data ratio at 1:9. We froze the parameters of the first six layers of the network to utilize the pre-trained model weights as shown in Table~\ref{tb2}. Additionally, we conducted a parallel experiment by training the AGCN model from scratch on the same dataset for a comparative analysis. We set the initial learning rate for this training at 0.05, which we reduced by a factor of ten at the $20_{th}$ and $40_{th}$ epochs. The training concluded after the $50_{th}$ epoch, while all other training parameters were consistent across experiments. The results of these experiments, which highlight the impacts and benefits of transfer learning, are documented in Table~\ref{tb3}.

\begin{table}[h]
    \caption{Comparison of classification accuracy using SkeFi framework on MM-Fi dataset. The * notation indicates that the SkeFi framework is not used.}
    \label{tb3}
    \centering
    \begin{tabular}{lccc}
    \hline
    Methods                              & RGB (\%)                              & mmWave (\%)                           & LiDAR (\%)                         \\ \hline
    STGCN~\cite{ST-GCN}                  & 73.50                                 & 44.75                                 & 38.74                                 \\ \hline
    AGCN~\cite{AGCN}                     & 90.99                                 & 58.09                                 & 65.39                                 \\
    AGC + MST~\cite{MSG3D}               & 92.84                                 & 60.21                                 & 65.04                                  \\
    AGC + ESP-MST                        & 93.97                                 & 62.22                                 & 66.54                        
                     \\ \hline
    TC-AGCN                              & 91.64                                 & 58.67                                 & 65.08                                 \\ 
    TC-AGC + MST                        & 93.02                                 & 62.78                                 & 65.31                                  \\
    \textbf{TC-AGC + ESP-MST}                    & \textbf{94.03} &  \textbf{62.98} & 65.02                       \\ \hline
    {*AGCN}~\cite{AGCN} & 57.67                                 & 41.89                                  & 20.63       \\ 
    {*TC-AGC + ESP-MST} & 62.61   & 46.23 & 49.49 \\ \hline
\end{tabular}
\end{table}

\begin{figure}[h]
\centering
\includegraphics[width=0.485\textwidth]{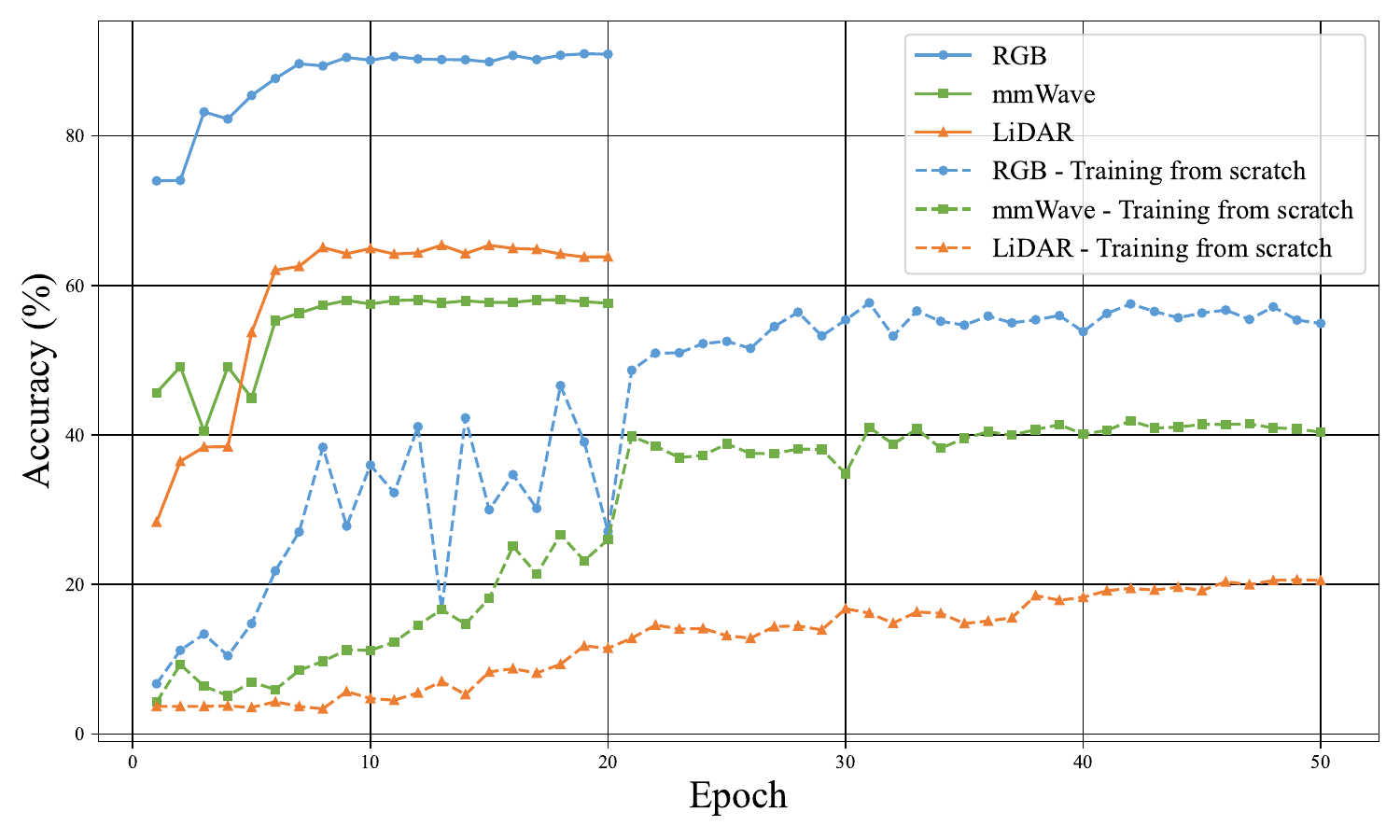}
\caption{The training process of AGCN in three modalities, where the dotted line represents training from scratch, and the solid line represents transfer learning.}
\label{fig5}
\end{figure}

In RGB and mmWave modalities, our proposed TC-AGCN demonstrates enhanced accuracy relative to AGCN's spatial modeling across three temporal modeling strategies. While TC-AGCN's performance on the Kinetics-Skeleton dataset is not notably preeminent, it consistently attains optimal accuracy within the modalities above. Training AGCN from scratch performs poorly in three modalities. Figure~\ref{fig5} visually represents the training dynamics of AGCN through three modes of transfer learning and direct training. These findings underscore that in scenarios with limited datasets, utilizing complex network architectures without sufficient data leads to ineffective learning of underlying data patterns. This not only impacts classification performance but also results in the inefficient use of computational resources.

\begin{figure}[h]
\centering
\includegraphics[width=0.485\textwidth]{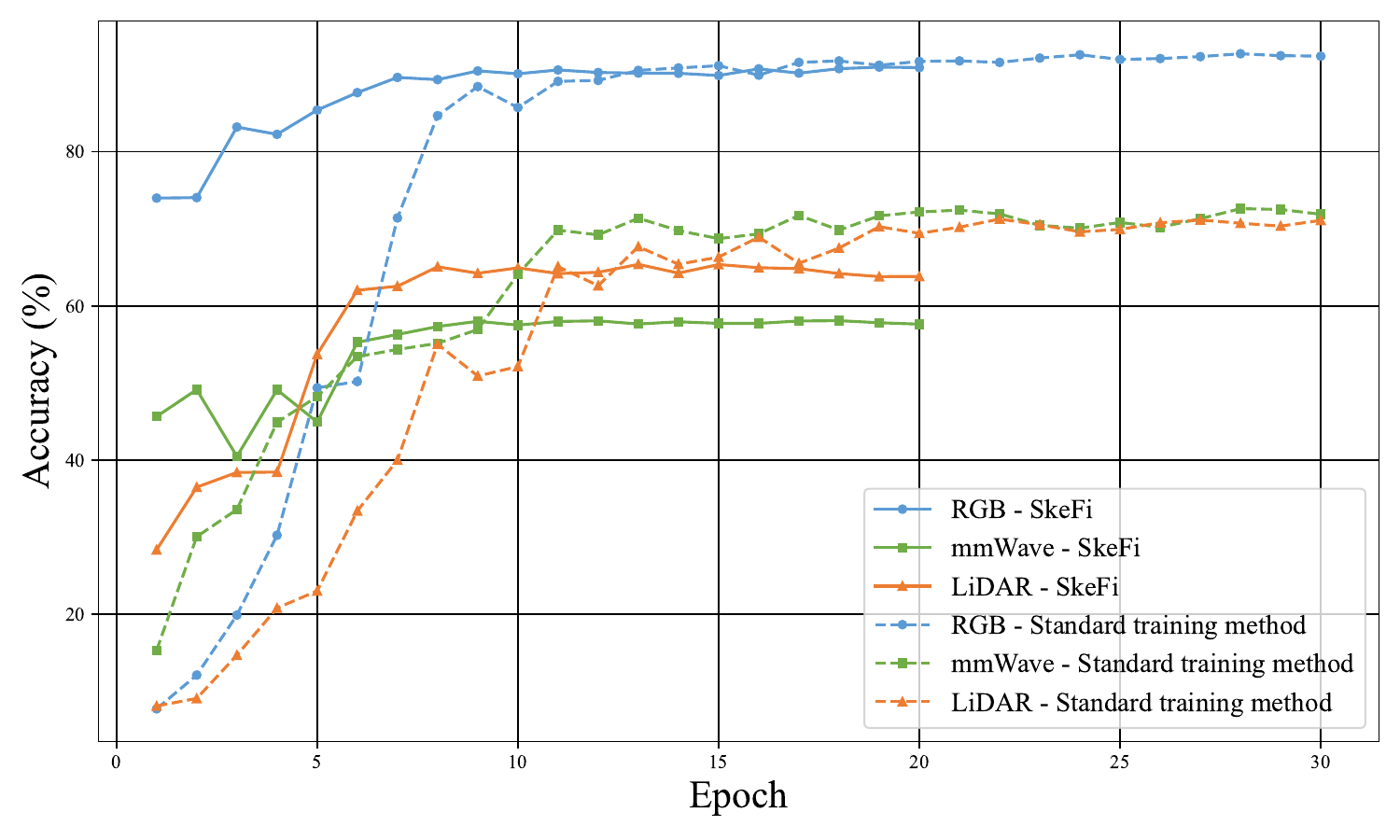}
\caption{The training process of AGCN in the standard training method and the SkeFi framework, where the dotted line represents the use of the standard training method and the solid line represents the use of the SkeFi framework.}
\label{fig6}
\end{figure}

Figure~\ref{fig6} illustrates the training process of AGCN using both standard training and the SkeFi framework. It is evident that under transfer learning with prior knowledge, the performance of the SkeFi framework in the RGB modality is comparable to the standard training method despite using only one subject as the training set for learning. For non-invasive data modalities such as mmWave and LiDAR, the performance difference between standard training and the SkeFi framework is 14\% and 6\%, respectively. These results demonstrate that our proposed SkeFi framework can perform well in action classification through transfer learning, even with limited training data.

\subsection{Other Factor Affecting Transfer Learning Performance}



\begin{figure}[h]
\centering
\includegraphics[width=0.485\textwidth]{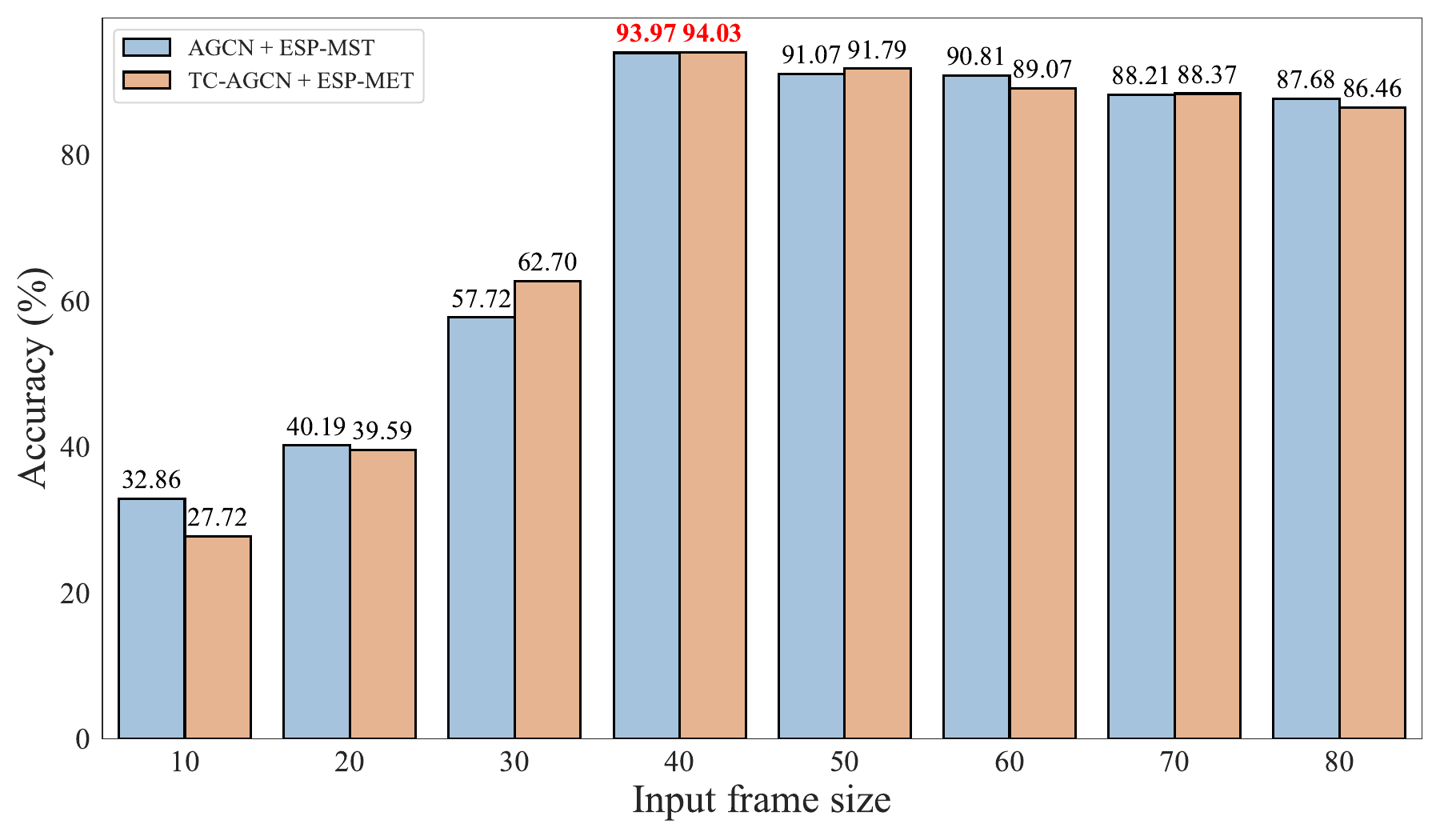}
\caption{The results of input different frame sizes in RGB modality.}
\label{fig8}
\end{figure}

\noindent\textbf{Number of frames for input data.} Upon examining the Kinetics-Skeleton and MM-Fi datasets, we found that each sample features the subject repeating the action multiple times. Given this shared characteristic, we follow~\cite{ST-GCN} and adopt the same data augmentation method as Kinetics-Skeleton for MM-Fi. Contrary to Kinetics-Skeleton, which chooses 150 frames at random from a total of 300 for training, our findings suggest that by appropriately expanding the number of frames in the original input data of MM-Fi ($3\times30\times18$), transfer learning accuracy can be markedly enhanced. Specifically, zero-padding the input data to extend it to 40 frames on both ends ($3\times40\times18$) yields the highest accuracy. The results for various frames are depicted in Figure~\ref{fig8}.

When contrasted with the utilization of the original input data, expanding the input data to encompass 40 frames enhances the accuracy by approximately 30\%. While accuracy remains commendable when the input data is expanded beyond 40 frames, it is worth noting that an increase in the number of frames also corresponds to a proportionate rise in computational resource consumption.

\section{Conclusion}
\label{sec:majhead}

This paper proposes SkeFi, a non-intrusive, data-driven cross-modal learning framework for action recognition, complemented by a temporally adaptive graph convolutional network (TC-AGCN) incorporating enhanced multi-scale temporal modeling. This method effectively tackles missing data frames and boosts temporal modeling's feature learning efficiency. Tested on the MM-Fi dataset, SkeFi outperforms baseline models trained from scratch, demonstrating significant computational efficiency. By comparing the SkeFi framework with standard training methods, it is evident that SkeFi can maintain a high level of classification performance even with limited training data. In addition, our findings indicate optimal fine-tuning and data-processing strategies substantially improve SkeFi's performance.

\section*{Acknowledgment}
This work is jointly supported by MOE Singapore Tier 1 Grant RG83/25, RS36/24 and a Start-up Grant from Nanyang Technological University.

\bibliographystyle{IEEEbib}
\bibliography{strings}

\begin{thebibliography}{10}

\bibitem{RNN}
Jun Liu, Amir Shahroudy, Dong Xu, and Gang Wang,
\newblock ``Spatio-temporal lstm with trust gates for 3d human action recognition,''
\newblock in {\em Computer Vision--ECCV 2016: 14th European Conference, Amsterdam, The Netherlands, October 11-14, 2016, Proceedings, Part III 14}. Springer, 2016, pp. 816--833.

\bibitem{CNN}
Hong Liu, Juanhui Tu, and Mengyuan Liu,
\newblock ``Two-stream 3d convolutional neural network for skeleton-based action recognition,''
\newblock {\em arXiv preprint arXiv:1705.08106}, 2017.

\bibitem{ST-GCN}
Sijie Yan, Yuanjun Xiong, and Dahua Lin,
\newblock ``Spatial temporal graph convolutional networks for skeleton-based action recognition,''
\newblock in {\em Proceedings of the AAAI conference on artificial intelligence}, 2018, vol.~32.

\bibitem{openpose}
Zhe Cao, Tomas Simon, Shih-En Wei, and Yaser Sheikh,
\newblock ``Realtime multi-person 2d pose estimation using part affinity fields,''
\newblock in {\em Proceedings of the IEEE conference on computer vision and pattern recognition}, 2017, pp. 7291--7299.

\bibitem{Nuitrack}
``Nuitrack full body skeletal tracking software - kinect replacement for android, windows, linux, ios, intel realsense, orbbec,'' \url{https://nuitrack.com/.}

\bibitem{posenet}
George Papandreou, Tyler Zhu, Nori Kanazawa, Alexander Toshev, Jonathan Tompson, Chris Bregler, and Kevin Murphy,
\newblock ``Towards accurate multi-person pose estimation in the wild,''
\newblock in {\em Proceedings of the IEEE conference on computer vision and pattern recognition}, 2017, pp. 4903--4911.

\bibitem{hr}
Guilherme Maeda, Marco Ewerton, Gerhard Neumann, Rudolf Lioutikov, and Jan Peters,
\newblock ``Phase estimation for fast action recognition and trajectory generation in human--robot collaboration,''
\newblock {\em The International Journal of Robotics Research}, vol. 36, no. 13-14, pp. 1579--1594, 2017.

\bibitem{VR}
Tamas Bates, Karinne Ramirez-Amaro, Tetsunari Inamura, and Gordon Cheng,
\newblock ``On-line simultaneous learning and recognition of everyday activities from virtual reality performances,''
\newblock in {\em 2017 IEEE/RSJ International Conference on Intelligent Robots and Systems (IROS)}. IEEE, 2017, pp. 3510--3515.

\bibitem{medical}
Amr Elkholy, Mohamed~E Hussein, Walid Gomaa, Dima Damen, and Emmanuel Saba,
\newblock ``Efficient and robust skeleton-based quality assessment and abnormality detection in human action performance,''
\newblock {\em IEEE journal of biomedical and health informatics}, vol. 24, no. 1, pp. 280--291, 2019.

\bibitem{yang2023sensefi}
Jianfei Yang, Xinyan Chen, Dazhuo Wang, Han Zou, Chris~Xiaoxuan Lu, Sumei Sun, and Lihua Xie,
\newblock ``Sensefi: A library and benchmark on deep-learning-empowered wifi human sensing,''
\newblock {\em Patterns}, vol. 4, no. 3, 2023.

\bibitem{Lidar}
Jialian Li, Jingyi Zhang, Zhiyong Wang, Siqi Shen, Chenglu Wen, Yuexin Ma, Lan Xu, Jingyi Yu, and Cheng Wang,
\newblock ``Lidarcap: Long-range marker-less 3d human motion capture with lidar point clouds,''
\newblock in {\em Proceedings of the IEEE/CVF Conference on Computer Vision and Pattern Recognition}, 2022, pp. 20502--20512.

\bibitem{mmWave}
Sizhe An and Umit~Y Ogras,
\newblock ``Mars: mmwave-based assistive rehabilitation system for smart healthcare,''
\newblock {\em ACM Transactions on Embedded Computing Systems (TECS)}, vol. 20, no. 5s, pp. 1--22, 2021.

\bibitem{wifi}
Jianfei Yang, Yunjiao Zhou, He~Huang, Han Zou, and Lihua Xie,
\newblock ``Metafi: Device-free pose estimation via commodity wifi for metaverse avatar simulation,''
\newblock in {\em 2022 IEEE 8th World Forum on Internet of Things (WF-IoT)}. IEEE, 2022, pp. 1--6.

\bibitem{nirmal2021deep}
Isura Nirmal, Abdelwahed Khamis, Mahbub Hassan, Wen Hu, and Xiaoqing Zhu,
\newblock ``Deep learning for radio-based human sensing: Recent advances and future directions,''
\newblock {\em IEEE Communications Surveys \& Tutorials}, vol. 23, no. 2, pp. 995--1019, 2021.

\bibitem{densepose}
R{\i}za~Alp G{\"u}ler, Natalia Neverova, and Iasonas Kokkinos,
\newblock ``Densepose: Dense human pose estimation in the wild,''
\newblock in {\em Proceedings of the IEEE conference on computer vision and pattern recognition}, 2018, pp. 7297--7306.

\bibitem{Resnet}
Kaiming He, Xiangyu Zhang, Shaoqing Ren, and Jian Sun,
\newblock ``Deep residual learning for image recognition,''
\newblock in {\em Proceedings of the IEEE conference on computer vision and pattern recognition}, 2016, pp. 770--778.

\bibitem{transfer}
Fuzhen Zhuang, Zhiyuan Qi, Keyu Duan, Dongbo Xi, Yongchun Zhu, Hengshu Zhu, Hui Xiong, and Qing He,
\newblock ``A comprehensive survey on transfer learning,''
\newblock {\em Proceedings of the IEEE}, vol. 109, no. 1, pp. 43--76, 2020.

\bibitem{transfer1}
Sinno~Jialin Pan and Qiang Yang,
\newblock ``A survey on transfer learning,''
\newblock {\em IEEE Transactions on knowledge and data engineering}, vol. 22, no. 10, pp. 1345--1359, 2009.

\bibitem{AGCN}
Lei Shi, Yifan Zhang, Jian Cheng, and Hanqing Lu,
\newblock ``Two-stream adaptive graph convolutional networks for skeleton-based action recognition,''
\newblock in {\em Proceedings of the IEEE/CVF conference on computer vision and pattern recognition}, 2019, pp. 12026--12035.

\bibitem{espnet}
Sachin Mehta, Mohammad Rastegari, Anat Caspi, Linda Shapiro, and Hannaneh Hajishirzi,
\newblock ``Espnet: Efficient spatial pyramid of dilated convolutions for semantic segmentation,''
\newblock in {\em Proceedings of the european conference on computer vision (ECCV)}, 2018, pp. 552--568.

\bibitem{MM-Fi}
Jianfei Yang, He~Huang, Yunjiao Zhou, Xinyan Chen, Yuecong Xu, Shenghai Yuan, Han Zou, Chris~Xiaoxuan Lu, and Lihua Xie,
\newblock ``Mm-fi: Multi-modal non-intrusive 4d human dataset for versatile wireless sensing,''
\newblock {\em arXiv preprint arXiv:2305.10345}, 2023.

\bibitem{RNN1}
Yong Du, Wei Wang, and Liang Wang,
\newblock ``Hierarchical recurrent neural network for skeleton based action recognition,''
\newblock in {\em Proceedings of the IEEE conference on computer vision and pattern recognition}, 2015, pp. 1110--1118.

\bibitem{MSG3D}
Ziyu Liu, Hongwen Zhang, Zhenghao Chen, Zhiyong Wang, and Wanli Ouyang,
\newblock ``Disentangling and unifying graph convolutions for skeleton-based action recognition,''
\newblock in {\em Proceedings of the IEEE/CVF conference on computer vision and pattern recognition}, 2020, pp. 143--152.

\bibitem{CTRGCN}
Yuxin Chen, Ziqi Zhang, Chunfeng Yuan, Bing Li, Ying Deng, and Weiming Hu,
\newblock ``Channel-wise topology refinement graph convolution for skeleton-based action recognition,''
\newblock in {\em Proceedings of the IEEE/CVF international conference on computer vision}, 2021, pp. 13359--13368.

\bibitem{DGSTGCN}
Haodong Duan, Jiaqi Wang, Kai Chen, and Dahua Lin,
\newblock ``Dg-stgcn: dynamic spatial-temporal modeling for skeleton-based action recognition,''
\newblock {\em arXiv preprint arXiv:2210.05895}, 2022.

\bibitem{PoseC3D}
Haodong Duan, Yue Zhao, Kai Chen, Dahua Lin, and Bo~Dai,
\newblock ``Revisiting skeleton-based action recognition,''
\newblock {\em arXiv preprint arXiv:2104.13586}, 2021.

\bibitem{ki}
Will Kay, Joao Carreira, Karen Simonyan, Brian Zhang, Chloe Hillier, Sudheendra Vijayanarasimhan, Fabio Viola, Tim Green, Trevor Back, Paul Natsev, et~al.,
\newblock ``The kinetics human action video dataset,''
\newblock {\em arXiv preprint arXiv:1705.06950}, 2017.

\bibitem{NTU}
Amir Shahroudy, Jun Liu, Tian-Tsong Ng, and Gang Wang,
\newblock ``Ntu rgb+ d: A large scale dataset for 3d human activity analysis,''
\newblock in {\em Proceedings of the IEEE conference on computer vision and pattern recognition}, 2016, pp. 1010--1019.

\bibitem{finegym}
Dian Shao, Yue Zhao, Bo~Dai, and Dahua Lin,
\newblock ``Finegym: A hierarchical video dataset for fine-grained action understanding,''
\newblock in {\em Proceedings of the IEEE/CVF conference on computer vision and pattern recognition}, 2020, pp. 2616--2625.

\bibitem{ucf101}
Khurram Soomro, Amir~Roshan Zamir, and Mubarak Shah,
\newblock ``Ucf101: A dataset of 101 human actions classes from videos in the wild,''
\newblock {\em arXiv preprint arXiv:1212.0402}, 2012.

\bibitem{WDHS1}
Kaixuan Chen, Dalin Zhang, Lina Yao, Bin Guo, Zhiwen Yu, and Yunhao Liu,
\newblock ``Deep learning for sensor-based human activity recognition: Overview, challenges, and opportunities,''
\newblock {\em ACM Computing Surveys (CSUR)}, vol. 54, no. 4, pp. 1--40, 2021.

\bibitem{WDHS2}
Chenning Li, Zhichao Cao, and Yunhao Liu,
\newblock ``Deep ai enabled ubiquitous wireless sensing: A survey,''
\newblock {\em ACM Computing Surveys (CSUR)}, vol. 54, no. 2, pp. 1--35, 2021.

\bibitem{WDHS3}
Athira Nambiar, Alexandre Bernardino, and Jacinto~C Nascimento,
\newblock ``Gait-based person re-identification: A survey,''
\newblock {\em ACM Computing Surveys (CSUR)}, vol. 52, no. 2, pp. 1--34, 2019.

\bibitem{wearable1}
Sumit Majumder, Tapas Mondal, and M~Jamal Deen,
\newblock ``Wearable sensors for remote health monitoring,''
\newblock {\em Sensors}, vol. 17, no. 1, pp. 130, 2017.

\bibitem{wearable2}
Ebrahim Nemati, M~Jamal Deen, and Tapas Mondal,
\newblock ``A wireless wearable ecg sensor for long-term applications,''
\newblock {\em IEEE Communications Magazine}, vol. 50, no. 1, pp. 36--43, 2012.

\bibitem{camera}
Cheng Zhang, Fan Yang, Gang Li, Qiang Zhai, Yi~Jiang, and Dong Xuan,
\newblock ``Mv-sports: a motion and vision sensor integration-based sports analysis system,''
\newblock in {\em IEEE INFOCOM 2018-IEEE Conference on Computer Communications}. IEEE, 2018, pp. 1070--1078.

\bibitem{XRF55}
Fei Wang, Yizhe Lv, Mengdie Zhu, Han Ding, and Jinsong Han,
\newblock ``Xrf55: A radio frequency dataset for human indoor action analysis,''
\newblock {\em Proceedings of the ACM on Interactive, Mobile, Wearable and Ubiquitous Technologies}, vol. 8, no. 1, pp. 1--34, 2024.

\bibitem{wifi2}
Qifan Pu, Sidhant Gupta, Shyamnath Gollakota, and Shwetak Patel,
\newblock ``Whole-home gesture recognition using wireless signals,''
\newblock in {\em Proceedings of the 19th annual international conference on Mobile computing \& networking}, 2013, pp. 27--38.

\bibitem{wifi10}
Hamada Rizk and Ahmed Elmogy,
\newblock ``Self-supervised wifi-based identity recognition in multi-user smart environments,''
\newblock {\em Sensors}, vol. 25, no. 10, pp. 3108, 2025.

\bibitem{mmWave2}
Zhicheng Yang, Parth~H Pathak, Yunze Zeng, Xixi Liran, and Prasant Mohapatra,
\newblock ``Monitoring vital signs using millimeter wave,''
\newblock in {\em Proceedings of the 17th ACM international symposium on mobile ad hoc networking and computing}, 2016, pp. 211--220.

\bibitem{mmWave3}
Haipeng Liu, Yuheng Wang, Anfu Zhou, Hanyue He, Wei Wang, Kunpeng Wang, Peilin Pan, Yixuan Lu, Liang Liu, and Huadong Ma,
\newblock ``Real-time arm gesture recognition in smart home scenarios via millimeter wave sensing,''
\newblock {\em Proceedings of the ACM on interactive, mobile, wearable and ubiquitous technologies}, vol. 4, no. 4, pp. 1--28, 2020.

\bibitem{mmWave4}
Zhen Meng, Song Fu, Jie Yan, Hongyuan Liang, Anfu Zhou, Shilin Zhu, Huadong Ma, Jianhua Liu, and Ning Yang,
\newblock ``Gait recognition for co-existing multiple people using millimeter wave sensing,''
\newblock in {\em Proceedings of the AAAI Conference on Artificial Intelligence}, 2020, vol.~34, pp. 849--856.

\bibitem{wang2023human}
Shuai Wang, Dongjiang Cao, Ruofeng Liu, Wenchao Jiang, Tianshun Yao, and Chris~Xiaoxuan Lu,
\newblock ``Human parsing with joint learning for dynamic mmwave radar point cloud,''
\newblock {\em Proceedings of the ACM on Interactive, Mobile, Wearable and Ubiquitous Technologies}, vol. 7, no. 1, pp. 1--22, 2023.

\bibitem{zou2019wifi}
Han Zou, Jianfei Yang, Hari Prasanna~Das, Huihan Liu, Yuxun Zhou, and Costas~J Spanos,
\newblock ``Wifi and vision multimodal learning for accurate and robust device-free human activity recognition,''
\newblock in {\em Proceedings of the IEEE/CVF conference on computer vision and pattern recognition workshops}, 2019, pp. 0--0.

\bibitem{gesture}
Han Zou, Jianfei Yang, Yuxun Zhou, Lihua Xie, and Costas~J Spanos,
\newblock ``Robust wifi-enabled device-free gesture recognition via unsupervised adversarial domain adaptation,''
\newblock in {\em 2018 27th International Conference on Computer Communication and Networks (ICCCN)}. IEEE, 2018, pp. 1--8.

\bibitem{wu2020fingerdraw}
Dan Wu, Ruiyang Gao, Youwei Zeng, Jinyi Liu, Leye Wang, Tao Gu, and Daqing Zhang,
\newblock ``Fingerdraw: Sub-wavelength level finger motion tracking with wifi signals,''
\newblock {\em Proceedings of the ACM on Interactive, Mobile, Wearable and Ubiquitous Technologies}, vol. 4, no. 1, pp. 1--27, 2020.

\bibitem{WiPose}
Wenjun Jiang, Hongfei Xue, Chenglin Miao, Shiyang Wang, Sen Lin, Chong Tian, Srinivasan Murali, Haochen Hu, Zhi Sun, and Lu~Su,
\newblock ``Towards 3d human pose construction using wifi,''
\newblock in {\em Proceedings of the 26th Annual International Conference on Mobile Computing and Networking}, 2020, pp. 1--14.

\bibitem{Winect}
Yili Ren, Zi~Wang, Sheng Tan, Yingying Chen, and Jie Yang,
\newblock ``Winect: 3d human pose tracking for free-form activity using commodity wifi,''
\newblock {\em Proceedings of the ACM on Interactive, Mobile, Wearable and Ubiquitous Technologies}, vol. 5, no. 4, pp. 1--29, 2021.

\bibitem{GoPose}
Yili Ren, Zi~Wang, Yichao Wang, Sheng Tan, Yingying Chen, and Jie Yang,
\newblock ``Gopose: 3d human pose estimation using wifi,''
\newblock {\em Proceedings of the ACM on Interactive, Mobile, Wearable and Ubiquitous Technologies}, vol. 6, no. 2, pp. 1--25, 2022.

\bibitem{c3t}
Abhi Kamboj, Anh~Duy Nguyen, and Minh Do,
\newblock ``C3t: Cross-modal transfer through time for human action recognition,''
\newblock {\em arXiv preprint arXiv:2407.16803}, 2024.

\bibitem{radardistill}
Geonho Bang, Kwangjin Choi, Jisong Kim, Dongsuk Kum, and Jun~Won Choi,
\newblock ``Radardistill: Boosting radar-based object detection performance via knowledge distillation from lidar features,''
\newblock in {\em Proceedings of the IEEE/CVF Conference on Computer Vision and Pattern Recognition}, 2024, pp. 15491--15500.

\bibitem{crkd}
Lingjun Zhao, Jingyu Song, and Katherine~A Skinner,
\newblock ``Crkd: Enhanced camera-radar object detection with cross-modality knowledge distillation,''
\newblock in {\em Proceedings of the IEEE/CVF Conference on Computer Vision and Pattern Recognition}, 2024, pp. 15470--15480.

\bibitem{ASGCN}
Maosen Li, Siheng Chen, Xu~Chen, Ya~Zhang, Yanfeng Wang, and Qi~Tian,
\newblock ``Actional-structural graph convolutional networks for skeleton-based action recognition,''
\newblock in {\em Proceedings of the IEEE/CVF conference on computer vision and pattern recognition}, 2019, pp. 3595--3603.

\bibitem{pytorch}
Adam Paszke, Sam Gross, Soumith Chintala, Gregory Chanan, Edward Yang, Zachary DeVito, Zeming Lin, Alban Desmaison, Luca Antiga, and Adam Lerer,
\newblock ``Automatic differentiation in pytorch,''
\newblock in {\em NIPS 2017 Workshop on Autodiff}, 2017.

\bibitem{SunXLW19}
Ke~Sun, Bin Xiao, Dong Liu, and Jingdong Wang,
\newblock ``Deep high-resolution representation learning for human pose estimation,''
\newblock in {\em CVPR}, 2019.

\bibitem{HrNet}
Jingdong Wang, Ke~Sun, Tianheng Cheng, Borui Jiang, Chaorui Deng, Yang Zhao, Dong Liu, Yadong Mu, Mingkui Tan, Xinggang Wang, Wenyu Liu, and Bin Xiao,
\newblock ``Deep high-resolution representation learning for visual recognition,''
\newblock {\em TPAMI}, 2019.

\bibitem{coco}
Tsung-Yi Lin, Michael Maire, Serge Belongie, James Hays, Pietro Perona, Deva Ramanan, Piotr Doll{\'a}r, and C~Lawrence Zitnick,
\newblock ``Microsoft coco: Common objects in context,''
\newblock in {\em Computer Vision--ECCV 2014: 13th European Conference, Zurich, Switzerland, September 6-12, 2014, Proceedings, Part V 13}. Springer, 2014, pp. 740--755.

\bibitem{freeze}
Jacob Devlin, Ming-Wei Chang, Kenton Lee, and Kristina Toutanova,
\newblock ``Bert: Pre-training of deep bidirectional transformers for language understanding,''
\newblock in {\em Proceedings of the 2019 conference of the North American chapter of the association for computational linguistics: human language technologies, volume 1 (long and short papers)}, 2019, pp. 4171--4186.

\bibitem{freeze1}
Jason Yosinski, Jeff Clune, Yoshua Bengio, and Hod Lipson,
\newblock ``How transferable are features in deep neural networks?,''
\newblock {\em Advances in neural information processing systems}, vol. 27, 2014.

\end{thebibliography}

\begin{IEEEbiography}
    [{\includegraphics[width=0.9in,clip,keepaspectratio]{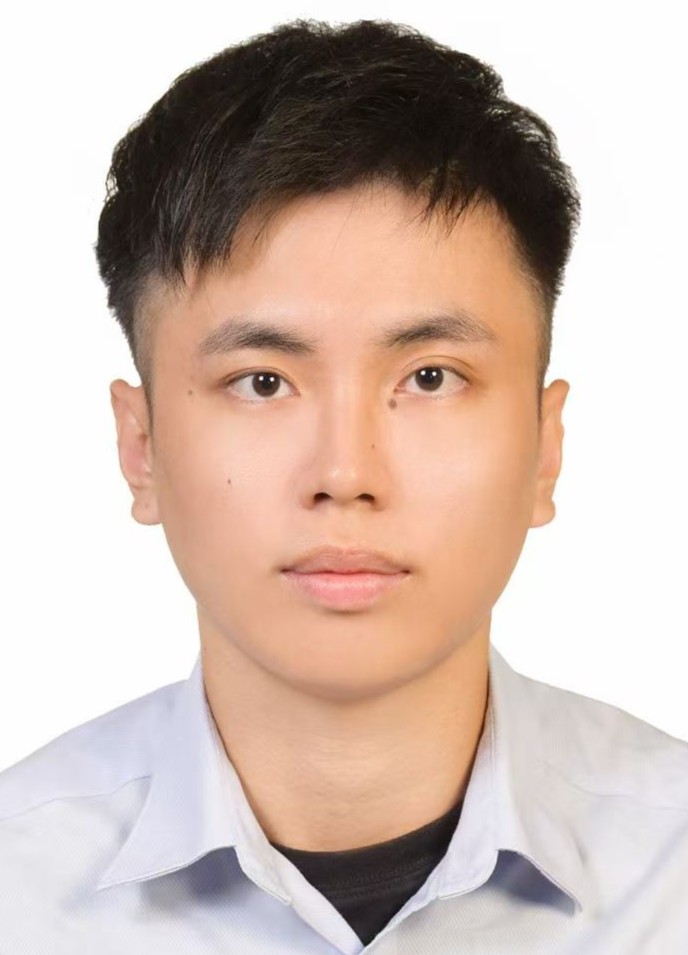}}]{Shunyu Huang}
    received the B.Eng. from the School of Artificial Intelligence and Automation at Huazhong University of Science and Technology in 2022 and the Msc. degree in Electrical and Electronic Engineering from Nanyang Technological University in 2024. Currently, he is a systems engineer at TSMC, developing systems for smart semiconductor manufacturing.
\end{IEEEbiography}

\begin{IEEEbiography}[{\includegraphics[width=0.9in,clip,keepaspectratio]{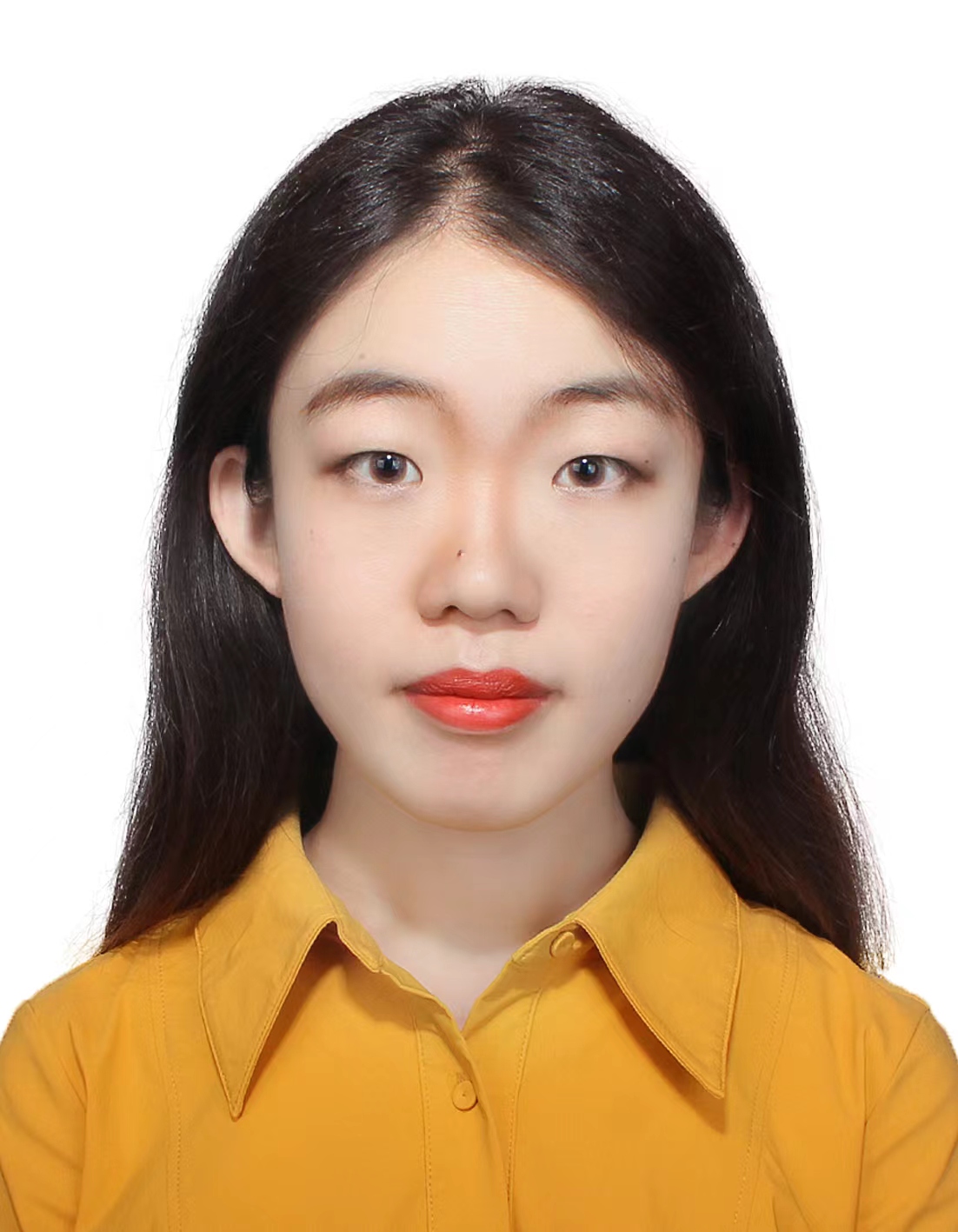}}]{Yunjiao Zhou}
	received the B.Eng. from the School of Information Science and Technology, Southwest Jiaotong in 2021, and the Msc. degree in Electrical and Electronic Engineering from Nanyang Technological University in 2022. She is currently pursuing a PhD degree in the Internet of Things laboratory, Nanyang Technological University. Her research interests include multi-modal learning for multi-sensor applications and human perception in smart home.
\end{IEEEbiography}

\begin{IEEEbiography}[{\includegraphics[width=0.9in,clip,keepaspectratio]{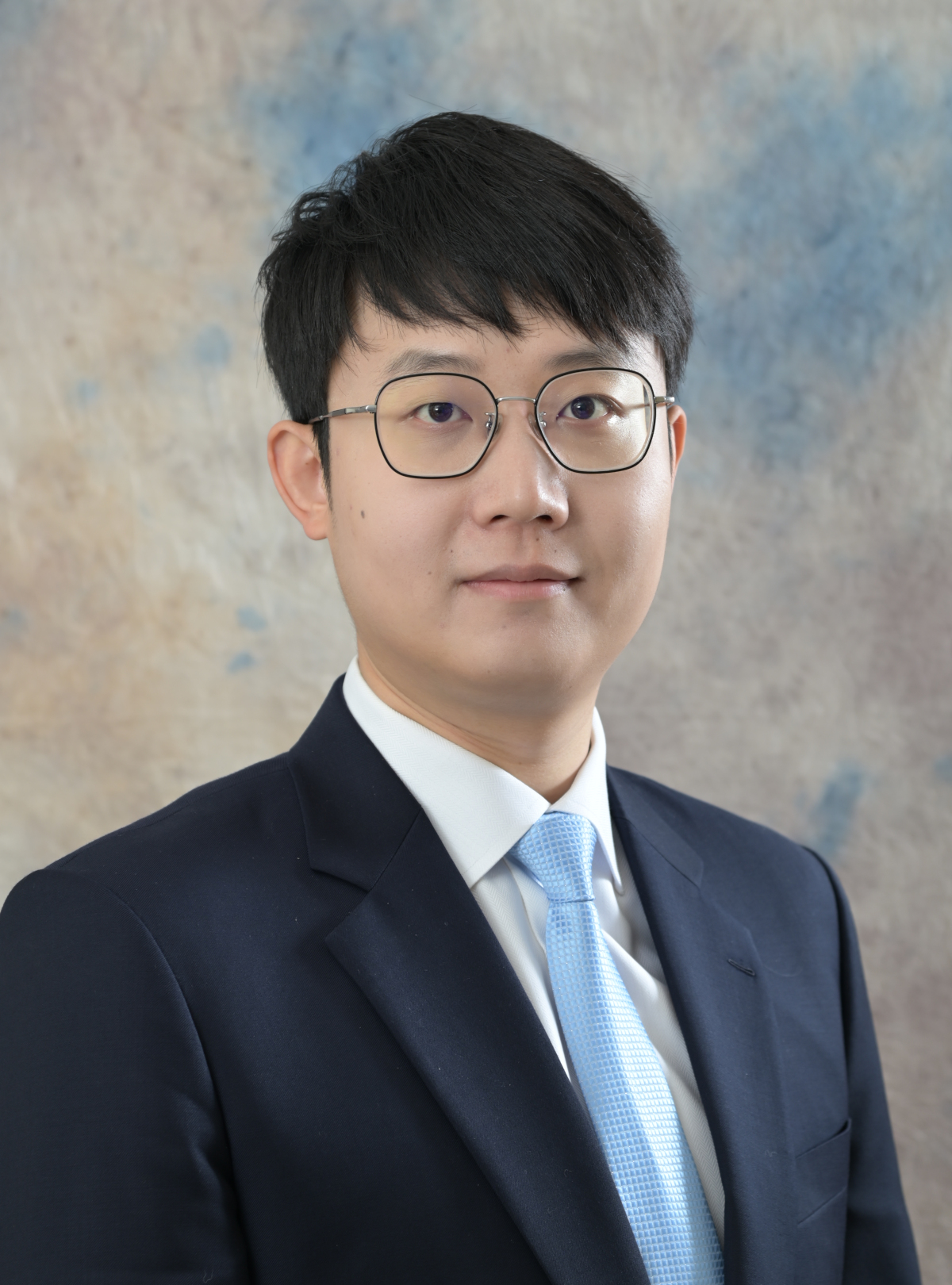}}]{Jianfei Yang}
(M'18–SM'25) is an Assistant Professor with the School of Mechanical and Aerospace Engineering, jointly appointed with the School of Electrical and Electronic Engineering at Nanyang Technological University (NTU), Singapore, where he leads the Multimodal Embodied AI and Robotic Systems (MARS) Lab. He received the B.Eng. degree from Sun Yat-sen University in 2016 and the Ph.D. degree from NTU in 2020 with the Best Thesis Award. After completing his Ph.D., he served as a Senior Research Engineer at the Berkeley Education Alliance for Research in Singapore (BEARS), University of California, Berkeley, and subsequently as a Presidential Postdoctoral Fellow at NTU from 2021 to 2023. In 2024, he was a Visiting Scholar at the University of Tokyo and Harvard University.

His research focuses on physical AI, exploring how artificial intelligence can empower physical systems—such as robots, IoT devices, and industrial platforms—to sense, understand, and interact with the real world. He has been recognized as a Forbes 30 Under 30 honoree (Asia, 2024) and has been listed among Stanford's World's Top 2\% Scientists since 2023. He has also won more than ten international AI challenges spanning computer vision, embodied AI, and interdisciplinary research areas.

\end{IEEEbiography}

\end{document}